\documentclass[conference]{IEEEtran}

\usepackage{amsmath,amsfonts,amsthm,amssymb,color}
\usepackage{pifont}
\usepackage{microtype}
\usepackage{graphicx}
\usepackage{subfigure}
\usepackage{booktabs} 
\usepackage{paralist}
\usepackage{amssymb}
\usepackage{amsmath}
\usepackage{empheq}
\usepackage{mdframed}
\usepackage[utf8]{inputenc}
\usepackage[english]{babel}
\usepackage{amsthm} 
\usepackage{xspace}
\usepackage{bm}
\usepackage{multirow}

\usepackage{algorithm}
\usepackage{algorithmic}

\usepackage{color, colortbl}
\usepackage{cellspace}
\usepackage{hyperref}

\def\BibTeX{{\rm B\kern-.05em{\sc i\kern-.025em b}\kern-.08em
    T\kern-.1667em\lower.7ex\hbox{E}\kern-.125emX}}

\newtheorem{problem}{Problem}

\theoremstyle{definition}
\newtheorem{definition}{Definition}[section]

\newcommand{\hide}[1]{}

\newcommand\scalemath[2]{\scalebox{#1}{\mbox{\ensuremath{\displaystyle #2}}}}

\newcommand{\la}[1]{{\color{red} LA: #1}}

\newcommand\sbullet[1][.75]{\mathbin{\vcenter{\hbox{\scalebox{#1}{$\bullet$}}}}}
\newcommand{\cmark}{\ding{51}}%

\makeatletter
\newcommand\footnoteref[1]{\protected@xdef\@thefnmark{\ref{#1}}\@footnotemark}
\makeatother

\newcommand{\cbit}{\begin{compactitem}}
\newcommand{\ceit}{\end{compactitem}}
\newcommand{\cben}{\begin{compactenum}}
\newcommand{\ceen}{\end{compactenum}}

\newcommand{\beq}{\begin{equation}}
	\newcommand{\eeq}{\end{equation}}

\newcommand{\bit}{\begin{itemize}}
	\newcommand{\eit}{\end{itemize}}
\newcommand{\ben}{\begin{enumerate}}
	\newcommand{\een}{\end{enumerate}}

\newcommand{\ny}{{Nystr\"{o}m}\xspace}

\newcounter{x}

\newcommand{\udr}{{\sc UDR}\xspace}
\newcommand{\mc}{{\sc MC}\xspace}
\newcommand{\hits}{{\sc HITS}\xspace}
\newcommand{\hitse}{{\sc HITS-ens}\xspace}

\newcommand{\pk}{{\sc PK}\xspace}
\newcommand{\wl}{{\sc WL}\xspace}
\newcommand{\gtovec}{{\sc g2vec}\xspace}

\newcommand{\MMD}{{\sc MMD}\xspace}

\newcommand{\mC}{\mathcal{C}}
\newcommand{\mG}{\mathcal{G}}
\newcommand{\mS}{\mathcal{S}}
\newcommand{\mH}{\mathcal{H}}
\newcommand{\mP}{\mathcal{P}}
\newcommand{\mN}{\mathcal{N}}
\newcommand{\mB}{\mathcal{B}}
\newcommand{\mW}{\mathcal{W}}

\newcommand{\bW}{\mathbf{W}}
\newcommand{\bK}{\mathbf{K}}
\newcommand{\bH}{\mathbf{H}}
\newcommand{\bU}{\mathbf{U}}
\newcommand{\bS}{\mathbf{\Sigma}}
\newcommand{\bV}{\mathbf{V}}
\newcommand{\bD}{\bm{\Lambda}}

\newcommand{\bx}{\mathbf{x}}
\newcommand{\by}{\mathbf{y}}
\newcommand{\bh}{\mathbf{h}}
\newcommand{\bhp}{\mathbf{h}^\prime}

\newcommand{\R}{\mathbb{R}}
\newcommand{\E}{\mathbb{E}}

\newcommand{\bq}{\mathbb{Q}}
\newcommand{\bp}{\mathbb{P}}
\newcommand{\bpi}{\mathbb{P}_i}
\newcommand{\bpip}{\mathbb{P}_{j}}
\newcommand{\hbpi}{\widehat{\mathbb{P}}_i}
\newcommand{\hbpip}{\widehat{\mathbb{P}}_{j}}

\newcommand{\method}{{\sc GLAM}\xspace}

\newcommand{\saurabh}[1]{{\color{green} Saurabh: #1}}

\begin{document}


\title{Graph Anomaly Detection with Unsupervised GNNs}

\author{
\IEEEauthorblockN{Lingxiao Zhao}
\IEEEauthorblockA{\textit{Carnegie Mellon University} \\
Pittsburgh, PA, USA \\
lingxiao@cmu.edu}
\and
\IEEEauthorblockN{Saurabh Sawlani}
\IEEEauthorblockA{\textit{SoundHound} \\
Berlin, Germany \\
saurabh.sawlani@gmail.com}
\and 
\IEEEauthorblockN{Arvind Srinivasan}
\IEEEauthorblockA{\textit{Carnegie Mellon University} \\
Pittsburgh, PA, USA \\
arvindsr@andrew.cmu.edu}
\and 
\IEEEauthorblockN{Leman Akoglu}
\IEEEauthorblockA{\textit{Carnegie Mellon University} \\
Pittsburgh, PA, USA \\
lakoglu@andrew.cmu.edu}
}

\maketitle

\begin{abstract}

Graph-based anomaly detection finds numerous applications in the real-world.
Thus, there exists extensive literature on the topic that has recently shifted toward deep detection models
due to advances in deep learning and graph neural networks (GNNs).
A vast majority of prior work  focuses on detecting node/edge/subgraph anomalies within a single graph,
with much less work on graph-level anomaly detection in a graph database. 
This work aims to fill two gaps in the literature: We
 		(1) design \method, an {end-to-end \textit{graph-level} anomaly detection} model based on GNNs, and
 		(2) focus on 
 		\textit{unsupervised model selection}, which is notoriously hard due to lack of any labels, yet especially critical for deep NN based models with
 		 a long list of hyperparameters.
\hide{
By utilizing GNNs, \method
 inherits two desirable properties that traditional solutions lack:
(a) it is end-to-end learnable, i.e. directly tied to an anomaly detection objective,  
and  (b) it can seamlessly handle
  graphs with a variety of properties (e.g. node labels/attributes). 
}
 Further, we propose a new pooling strategy  for graph-level embedding, called MMD-pooling, that is geared toward detecting \textit{distribution} anomalies which has not been considered before.
Through extensive experiments on 15 real-world datasets, we show that ($i$) \method outperforms 
 node-level and two-stage (i.e. not end-to-end) baselines, 
 and ($ii$) model selection picks a significantly more effective model than expectation (i.e. average) --without using any labels-- among candidates with otherwise large variation in performance.
\end{abstract}


\vspace{-0.05in}
\section{Introduction}
\label{sec:intro}

Given a \textit{collection} of graphs, possibly with weighted edges, and labeled or multi-attributed nodes,
how can we identify the anomalous graphs that stand out from the majority?
Graph-level anomaly detection, different from detecting anomalies in a \textit{single} graph, aims to discover unusual graphs among \textit{multiple} graphs in a (graph) {database}.
The problem applies to many real-world domains, where each graph may capture a  chemical compound \cite{journals/jbcb/JiangM18}, human pose \cite{conf/cvpr/MarkovitzSFZA20},  cargo shipment \cite{eberle2009identifying}, system-call \cite{conf/kdd/ManzoorMA16}, command flow \cite{conf/sdm/LiuYYHY05}, information cascade \cite{monti2019fake}, etc.




Detecting graph-level anomalies is a challenging problem.
Dependencies in structured data render traditional outlier detectors for point-cloud data inapplicable.
Other properties add to the representational complexity, such as edge weights, (discrete) node labels, or multiple (numeric) node attributes. 
There are also challenges unique to a graph database, namely size variability of graphs and non-correspondence/un-alignment of nodes among graphs (unlike e.g. in time-evolving graphs).


Graph-based anomaly detection has been studied in the literature  
\cite{AkogluTK15}.
However,
majority of the work focuses on (node, edge, subgraph) anomalies  within a \textit{single} graph, most often in plain graphs \cite{conf/kdd/IdeK04,conf/icdm/SunQCF05,conf/pakdd/AkogluMF10,conf/sdm/TongL11,conf/kdd/HooiSBSSF16}, and less often in labeled \cite{conf/kdd/NobleC03} or
 attributed graphs \cite{conf/kdd/PerozziASM14,journals/corr/ShahBHAGMKF15,conf/sdm/PerozziA16}.
Prior work on graph-level anomaly detection is much sparser, 
majority of which are traditional substructure mining based techniques \cite{conf/kdd/NobleC03,conf/kdd/ManzoorMA16,conf/kdd/EswaranFGM18}. These
 do not easily generalize to graphs with complex properties; e.g. SpotLight \cite{conf/kdd/EswaranFGM18} cannot accommodate node labels or attributes, Subdue \cite{conf/kdd/NobleC03} and StreamSpot \cite{conf/kdd/ManzoorMA16} cannot handle weighted edges or multi-attributed nodes.


With recent advances in deep learning, focus has shifted toward deep neural network based anomaly detection models (see surveys,  \cite{pang2021deep,journals/corr/abs-2009-11732,ma2021comprehensive}). Graph neural networks (GNNs) have been used for graph anomaly detection \cite{conf/kdd/YuCAZCW18,conf/sdm/DingLBL19,ocgnn20,conf/cikm/0003DYJW020,zhao2021using}, most of which have been limited to \textit{node-level} detection with one exception \cite{zhao2021using} that focused on graph-level anomaly evaluation. Similarly, graph representation learning has mainly focused on \textit{node-level} embeddings \cite{deepwalk2014,conf/nips/HamiltonYL17,hamilton2017representation}.
There exist graph-level embedding techniques based on NNs \cite{pmlr-v48-niepert16,NarayananCVCLJ17,conf/iclr/XuHLJ19} or graph kernels \cite{shervashidze2011weisfeiler,yanardag2015deep,neumann2016propagation,journals/corr/abs-1904-12218},
however those do not tackle the anomaly detection problem \textit{per se}.
These graph embeddings can be input to point-cloud outlier detectors; in such a two-pronged approach the first step
(graph embedding) is disconnected from the final goal (anomaly detection).
A conceptual comparison of related work is given in Table \ref{tab:related}.

GNN-based techniques are popular thanks to their expressiveness (layers of learnable parameters) and flexibility (ability to  handle graphs with complex properties). 
However, they rely on many hyperparameters (HPs) that influence their performance, such as number of hidden layers/units and epochs, drop-out/weight decay/learning rates, to name a few  \cite{journals/corr/abs-1811-05868}.
Careful tuning of the HPs, namely the model selection task, is therefore of utmost importance.
On the other hand, model selection is challenging for 
\textit{unsupervised} anomaly detection, in the absence of any labeled data.
To our surprise, recent work 
on deep outlier models have not recognized or emphasized this challenge \cite{chalapathy2019learning,conf/sdm/DingLBL19,pang2021deep}. 
Most use a fixed configuration, while some use labeled hold-out (validation) data for model tuning \cite{journals/corr/abs-2009-11732,ocgnn20}, 
which is prohibitive in fully unsupervised settings.

\begin{table*}
	\caption{Comparison of related work in terms of desired properties for graph-level anomaly detection. Also see Sec. \ref{sec:related}.}
	\label{tab:related}
	\vspace{-0.1in}	
	\scalebox{0.9}{
		\begin{tabular}{l|c|ccc|ccc|ccc|c}
			\toprule
			& 
			\cite{chalapathy2019learning,journals/corr/abs-2009-11732} &
			\cite{deepwalk2014,conf/nips/HamiltonYL17,conf/cikm/0003DYJW020,conf/kdd/YuCAZCW18} & 
			\cite{ocgnn20,conf/sdm/DingLBL19} &  
			\cite{KumagaiIF20} &  
			\cite{conf/kdd/NobleC03}  & 
			\cite{conf/kdd/ManzoorMA16} &     
			\cite{conf/kdd/EswaranFGM18} & 
			
			\cite{pmlr-v48-niepert16,conf/iclr/XuHLJ19} & 
			\cite{neumann2016propagation} &  
			\cite{NarayananCVCLJ17,shervashidze2011weisfeiler} & 
			this  \\	
			\textbf{Desired Properties}  & point-cloud 	& 	\multicolumn{3}{c|}{node-level} & \multicolumn{3}{c|}{traditional}& \multicolumn{3}{c|}{graph embedding} & paper \\
			\midrule
			end-to-end anomaly detection  & \cmark& & \cmark	&	 \cmark&		&		&	& &		 & &  \cmark\\\hline
			unsupervised (vs. (semi-)supervised)		&  \cmark&	\cmark	& \cmark	&	&	\cmark & \cmark& \cmark&	&  \cmark	& \cmark&	\cmark \\\hline
			graph-structured data (vs. point-cloud) 	&  	  & \cmark & \cmark	&	 \cmark&\cmark&	\cmark	&\cmark &	\cmark &\cmark &\cmark &	  \cmark\\\hline
			graph-level detection (vs. node/edge-level) &  	  	& &		&	& \cmark&	\cmark	&\cmark&\cmark	& \cmark &  \cmark&	 \cmark\\\hline
			graph embedding 			& 	& & 	&	&  &	 \cmark& \cmark	&\cmark &	\cmark	& \cmark&\cmark	\\\hline
			handle labeled nodes 							& 	& \cmark &	 \cmark &	 \cmark &\cmark&	\cmark & &	\cmark& \cmark &\cmark&	 	 \cmark\\\hline
			handle multi-attributed nodes						&  			&\cmark & \cmark& 	 \cmark&& & &	\cmark&	\cmark &	    	& 	 \cmark\\\hline
			handle  weighted edges	&  & \cmark	& \cmark	&\cmark	& 	& & \cmark	&\cmark	&\cmark	&	& 	\cmark \\\hline
			unsupervised model selection					&  &	&	&	&	& & 	& 		& 	&	&\cmark \\	
			\bottomrule
		\end{tabular}
	}
\vspace{-0.15in}	
\end{table*}


Accordingly, this work fills two gaps in the literature: We (1) design a GNN-based end-to-end model called \method to address the graph-level anomaly detection problem, and (2) address the unsupervised model selection task, that is, effectively select the hyperparameters of \method \textit{without using any labels}.
Further, we specify two different types of ($i.$ point and $ii.$ distribution) graph anomalies, design a novel pooling strategy for the latter, and demonstrate that they complement each other, thereby improving overall detection rate.  
Our contributions are summarized as follows.


\cbit
\item {\bf Deep Graph-level Anomaly Detection:} 
We propose \method, 
a novel \underline{G}raph-\underline{L}evel \underline{A}nomaly detection \underline{M}odel based on GNNs.
It embeds graphs in two ways:  by \textit{mean-pooling} and our newly proposed \textit{MMD-pooling} to detect point and distribution anomalies, respectively.
The latter 
treats each graph as a \textit{set} of its node embeddings, and helps identify complementary anomalies.
\hide{
In a nutshell it consists of three parts: 
($i$) node embedding, where each graph is ``flattened'' into a set of vectors,
($ii$) graph embedding, where each graph is represented either by \textit{mean-pooling} or as our newly proposed \textit{MMD-pooling} that captures the \textit{distribution} of its node embeddings,
and finally ($iii$) a one-class classification based objective at the output layer.
In contrast to all prior work, we treat a graph as a \textit{set} (of node embeddings), where graph comparison is done via  
 distribution kernels.
}

\item {\bf Desirable properties:}
As a GNN model \method 
inherits
two key properties: ($i$) expressiveness; multiple parameterized layers enable learnable embeddings that are directly tied to an anomaly detection objective, and 
($ii$) flexibility; various types of input graphs (directed, weighted, node-labeled/attributed) can be handled seamlessly. 



\item {\bf Unsupervised Anomaly Model Selection:}
Besides various advantages, \method also inherits a list of hyperparameters that require tuning for effective performance. Thus, we 
systematically address the unsupervised model selection (UMS) problem. 
To our knowledge, we are the first to recognize the importance and explicitly address UMS as part of deep learning based anomaly detection.


\item {\bf Effectiveness:}
Through experiments on 15 real-world graph databases, we show that ($i$) the effectiveness of \method against 8 GNN-based (two-stage and node-level) baselines,
($ii$) the ability of our UMS component to pick a model with superior performance as compared to a model with fixed configuration, 
and ($iii$)  the contributing factors behind \method through various ablation studies.

\ceit

To foster progress in graph-level anomaly detection and the related unsupervised model selection (UMS) problem, we open-source all code and data at \url{https://github.com/sawlani/GLAM}. 

\section{Preliminaries \& Background}
\label{sec:prelim}

We consider anomaly detection in a graph database $\mG=\{G_1$$=$$(V_1,E_1),$ $\ldots,  G_N$$=$$(V_N, E_N)\}$, containing graphs with labeled or attributed nodes, which we define as follows.

\begin{definition}[(Node-)Labeled Graph]
A labeled graph $G = (V,E)$ is endowed with a function $f : V \rightarrow \Sigma$ that assigns labels to the nodes from a discrete set of labels $\Sigma$.
\end{definition}


\vspace{-0.1in}
\begin{definition}[(Node-)Attributed Graph]
	An attributed graph $G = (V, E)$ is endowed with a function $g : V \rightarrow \R^d$ that assigns real-valued vectors to the nodes of the graph.
\end{definition}

For a graph $G=(V,E)$, let $X_v$, $v\in V$, denote the node vectors capturing the attributes, one-hot node labels or degrees respectively for attributed, labeled and plain graphs.
 Then, the graph anomaly detection problem is stated (informally) as:


\begin{problem}[Graph-level anomaly detection (GLAD)]
	\label{prob:glad}
\underline{Given} an unlabeled graph database $\mG=\{G_i=(V_i,E_i)\}_{i=1}^N$ containing $N$ unordered, node-labeled or node-attributed graphs; \underline{Identify} the unusual graphs that differ significantly from the majority of graphs in $\mG$.	
\end{problem}

Note that while all the graphs share the functions $f$ or $g$, they may be of different sizes, varying in the number of nodes and edges, where $|V_i|=n_i$ and $|E_i|=m_i$ for $i=\{1,\ldots,N\}$. Moreover, their nodes 
are \textit{not} necessarily in correspondence or otherwise the alignment is unknown (e.g. command flow graphs of two different software).
As such, the anomaly detection problem is different from, e.g., event detection for time-varying graphs, where the node correspondence is typically known and there exists a meaningful ordering among the graphs.

Importantly, GLAD comes bundled with an associated problem: unsupervised model selection.
As many models (especially deep NNs) exhibit a list of hyperparameters and the performance is sensitive to the choice of their values \cite{Campos2016}, it is critical to address GLAD with a built-in UMS solution.

\begin{problem}[Unsupervised Model Selection (UMS)]
	\label{prob:ums}
Consider an anomaly detection model $M(\boldsymbol{\Theta})$ for GLAD, where $\boldsymbol{\Theta}$ denotes the set of hyperparameters. \underline{Given} a new unlabeled dataset $\mG$, \underline{Select} an effective model -- without using any labels -- among candidates $\{M({\Theta}_1), M({\Theta}_2), \dots\}$  induced by different $\boldsymbol{\Theta}$ configurations/values. 
\end{problem}

\subsection{Graph Representation Learning}
{\bf Motivation.~} We seek to design a flexible anomaly detection approach that can accommodate graphs with a variety of characteristics. In particular, the detector should seamlessly apply to plain, node-labeled, as well as node-attributed graphs. 
The graph edges could also be weighted or directed.

Traditional solutions \cite{conf/kdd/NobleC03,conf/kdd/ManzoorMA16,conf/kdd/EswaranFGM18} struggle to generalize to all these settings, in fact, we are not aware of any that can handle multi-attributed nodes.
In contrast, the revolutionary neural network based graph representation learning techniques can accommodate graphs of any nature. Therefore, we build on graph neural networks (GNNs) for graph embedding.


\noindent
{\bf How-to.~} Modern GNNs employ a neighborhood aggregation
procedure, where the representation of a node is updated recursively by aggregating representations
of its neighbors over iterations.  
A node $v$'s representation after $l$ iterations, denoted $\bh_v^{(l)}$, is given as 
\beq
\label{eq:noderep}
\scalemath{0.9}{
\bh_v^{(l)} = \text{\sf COMBINE}^{(l)} \big(  \bh_v^{(l-1)},\; 
\text{\sf AGG}^{(l)}   \big(  \{ \bh_u^{(l-1)}: u\in \mN(v) \} \big)
  \big)
}
\eeq
where $\bh_v^{(0)}=X_v$ and $\mN(v)$ is $v$'s neighbor set.
The choice of the {\sf AGG} and {\sf COMBINE} functions differs among GNNs. 

Graph-level embedding is obtained by the so-called \textsf{READOUT} function that aggregates node embeddings after the final iteration $L$ to
obtain a representation $h_G$ for the entire graph $G=(V,E)$, i.e.,
\beq
\label{eq:graphrep}
\bh_G = \text{\sf READOUT} \big(  \{ \bh_v^{(L)} \; \vert \; v\in V \} \big)
\eeq
where 
{\sf READOUT} is a permutation-invariant function such as sum, mean (to account for size differences), or maximum. 

\subsection{Maximum Mean Discrepancy (MMD)}

{\bf Motivation.~}
GNN models produce node embeddings, each capturing 
the structural
information within its local, $L$-hop graph neighborhood.
``Blending'' those individual pieces by simple averaging to obtain graph-level embedding may be too na\"ive, 
obscuring key differences between graphs.

Alternatively, 
we regard each graph as a
\textit{collection} of node embeddings, and treat it as a \textit{distribution} in the (node) representation space, rather than a single point.
For anomaly detection, then, we aim to identify graphs whose distribution of nodes is significantly different from others.

\noindent
{\bf How-to.~} 
To compare two distributions, we use their kernel mean embedding within a reproducing kernel Hilbert space (RKHS) \cite{journals/jmlr/SriperumbudurGFSL10}.
The key idea is to map probability distributions injectively (i.e., one-to-one) into a RKHS with an associated kernel such that the distributions can be compared based on their metric distance in the RKHS. 
Such a metric is referred to as Maximum Mean Discrepancy (MMD) \cite{journals/jmlr/GrettonBRSS12}.

MMD defines a mapping from a class of probability distributions, denoted $\mP$, to the RKHS (of functions), denoted $\mH$, with a kernel $k(\cdot,\cdot)$ such that each $\bp\in \mP$ is mapped to a function $\mu_{\bp}: \R^d \mapsto \R$ in $\mH$, given as follows.
\begin{definition}[Kernel Embedding]
	\label{eq:meanembed}
	The \textit{kernel embedding of a probability distribution} $\bp \in \mP$ is given by the mapping	
\beq
	\label{eq:meanembedeq}
\hspace{-0.05in}
\scalemath{0.9}{
\hspace{-0.1in} \bp \mapsto \mu_{\bp} (\cdot) = \E_{\bp}[k(\cdot, X)] = \int_{\bx \in \R^d} k(\cdot, \bx)d\bp(\bx) \;, \text{ where } X\sim \bp 
}
\eeq
\end{definition}
The embedding $\mu_{\bp} (\cdot)$ acts as a representative function 
 in $\mH$ for $\bp$.
An injective embedding is desirable, such that each $\bp\in \mP$ is mapped to a \textit{unique} element $\mu_{\bp} \in \mH$, and can be obtained via the choice of a characteristic kernel for $k(\cdot,\cdot)$, such as Gaussian or Laplace kernels.
It follows from the reproducing property of $\mH$ that 
$\E_{\bp}[f] = \langle \mu_{\bp}, f  \rangle_{\mH} $ for all $f \in \mH$.
 
MMD, associated with function $\MMD: \mP \times \mP \mapsto \R$, is defined as the distance between the mean embeddings $\mu_{\bp}$ and 
$\mu_{\bq}$ of two distributions $\bp \in \mP$ and $\bq \in \mP$, that is,
\beq
\label{eq:mmd}
\MMD(\bp,\bq) = \| \mu_{\bp}-\mu_{\bq} \|_{\mH} \;.
\eeq

Due to the reproducing property, it can be shown that
\begin{equation}
\label{eq:mmdsq}
\scalemath{0.9}{
\hspace{-0.1in}
\MMD^2(\bp,\bq)  = \E_{\bx,\bx'}[k(\bx,\bx')] + \E_{\by,\by'}[k(\by,\by')] 
 - 2 \E_{\bx,\by}[k(\bx,\by)]
}
\end{equation}
where $\bx, \bx'$ and $\by, \by'$ are distributed according to $\bp$ and $\bq$, respectively. 


Note that one can estimate Eq. \eqref{eq:mmdsq} empirically, provided samples drawn from $\bp$ and $\bq$.
In other words, we can
easily and directly compare two distributions based on samples alone, \textit{without estimating any probability density functions} as an intermediate step.


\subsection{One-Class Classification}

{\bf Motivation.~} 
We seek to design an end-to-end approach that is
tied directly to an anomaly detection objective. 
The reasons are two-fold:
First, having an optimization criterion makes the task concrete and explicit. In comparison, most existing detectors are procedural or measure-based, based on distances \cite{ramaswamy2000efficient,journals/vldb/KnorrNT00}, densities \cite{breunig2000lof,tang2002enhancing,goldstein2012histogram}, etc., which perform intuitive computations that are otherwise not optimized.
Second,  end-to-end anomaly detection enables learnable graph representations through parameterized layers, which require a loss function for training purposes. This is in contrast to unsupervised graph embedding \cite{NarayananCVCLJ17,shervashidze2011weisfeiler,neumann2016propagation}, where the task is isolated from the end goal (anomaly detection).

\noindent
{\bf How-to.~}
In anomaly detection applications, the input examples are often unlabeled but belong mostly to the inlier category. 
As such, anomaly detection can be cast as a one-class classification problem to model this majority class.

A popular technique 
is the one-class SVM (OCSVM) \cite{ScholkopfPSSW01}, which 
estimates a max-margin hyperplane separating the training points from the origin, in effect treating the origin as the only negative data point.
Another similar approach, called Support Vector Data Description (SVDD) \cite{TaxD04}, aims to find a small-radius 
hypersphere that encloses majority of training data.
These work in the original or the kernelized feature space.

%

Most recently, deep neural network (DNN) counterparts have been developed for one-class classification.
Most of these, 
e.g. Deep-SVDD \cite{RuffGDSVBMK18} and DROCC \cite{goyal2020drocc},
capitalize on the ability of DNNs to learn latent representations, which are  optimized through a one-class objective at the output layer.
For a survey of deep one-class anomaly detection, see \cite{pang2021deep,journals/corr/abs-2009-11732}.



\section{Proposed Method: \method}
\label{sec:method}


We introduce \method for graph-level anomaly detection (Problem \ref{prob:glad}), \textit{built-in} with UMS (Problem 2).
\method first generates graph-level representations with GNNs using both mean and MMD-pooling (Sec. \ref{ssec:rep}), on top of which it employs an anomaly detection objective (Sec. \ref{ssec:anomaly}), and finally performs UMS (Sec. \ref{sec:supgpca}).

%
%
%
%
%
%

\begin{figure*}[!ht]
\vspace{-0.1in}
    \includegraphics[width=0.62\linewidth]{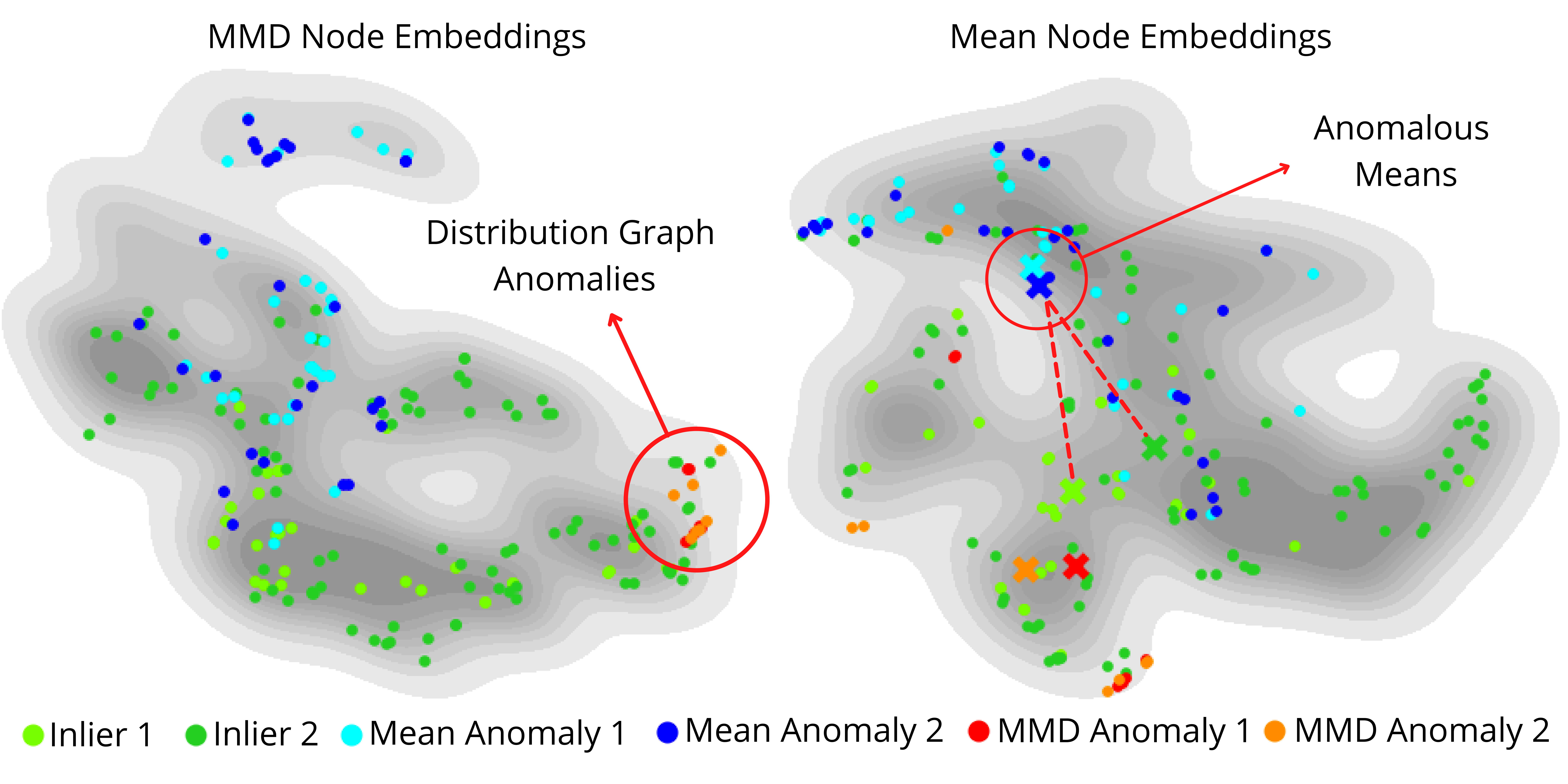}
    \includegraphics[width=0.35\linewidth]{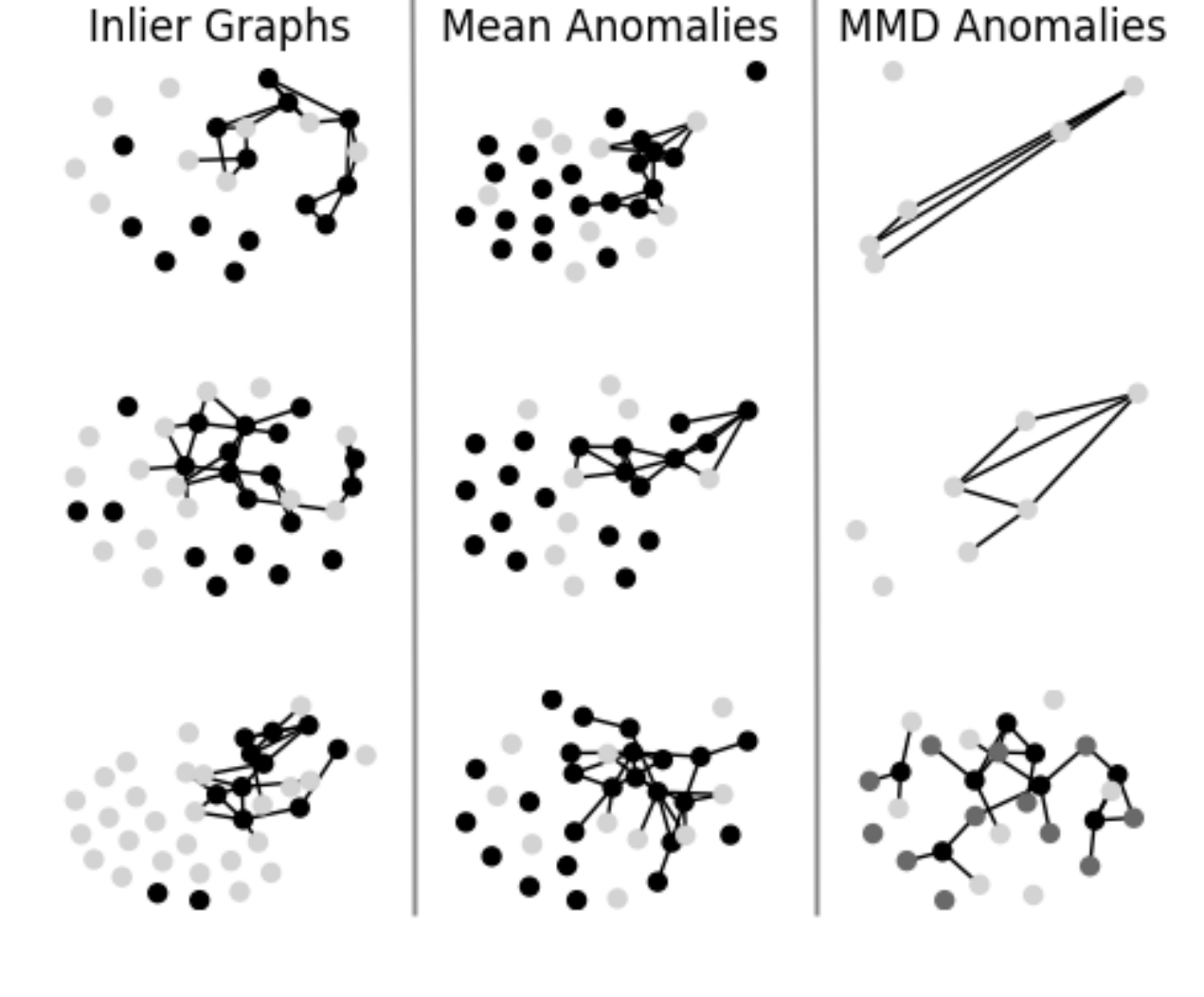}
    
	\vspace{-0.15in}
	\caption{Node embedding space by (left) MMD- and (middle) Mean-pooling in \textsc{Proteins}. Heatmap (gray) reflects the background distribution; i.e. density of all node embeddings across $\mG$. Symbols w/ same color are nodes of the same graph (see legend). 
		Notice that MMD is complementary to Mean: ($i$) Distribution of node-embeddings (red and orange dots) for MMD-anomalies (on left) differs significantly from background, yet (in middle) Mean-pooling misses them as their means (red\&orange crosses) are close to the means of inlier graphs (light/dark green crosses). ($ii$) On the other hand, node-embeddings of Mean-anomalies blend well into background distribution (on left) while (in middle) their means (light/dark blue crosses) are far away from those of inliers (light/dark green crosses). 
		 (right) E.g. inliers, and top Mean- and MMD-anomalous graphs. MMD anomalies have distinct distributions, w.r.t. graph structure and/or node labels.
		\label{fig:background}}
	\vspace{-0.2in}
\end{figure*}

\vspace{-0.05in}
\subsection{Graph-level Representation}
\label{ssec:rep}

The first step of \method is ``flattening'' each graph to a set of node embeddings (i.e. vectors).
To this end, 
we employ the GIN model that was shown to be
one of the most expressive among a large class of MPNNs \cite{conf/iclr/XuHLJ19}.
To model the \text{\sf COMBINE} and \text{\sf AGGREGATE} functions in Eq. \eqref{eq:noderep},
GIN employs multi-layer perceptrons (MLPs), with learnable parameters, for their injectiveness property. 
As MLPs can represent the composition of functions, the update equation of node embeddings at layer $l$ is  written as
\beq
\label{eq:noderepgin}
\scalemath{0.9}{
\bh_v^{(l)} = \text{\sf MLP}^{(l)} \big( (1+\epsilon^{(l)}) \cdot \bh_v^{(l-1)}
+
 \sum_{u\in \mN(v)}  \bh_u^{(l-1)}
\big) \;.
}
\eeq
\vspace{-0.1in}

Upon node embedding (by $L$-layer GIN), each graph $G_i\in \mG$ can be regarded as a set 
$\mS_i = \{\bh_{1,i}\;, \;\bh_{2,i}\;, \ldots, \;\bh_{n_i, i}\}$ (superscript $^{(L)}$'s dropped for simplicity) with cardinality $n_i=|V_i|$ where $\bh_{v,i} \in \R^{d'}$  is the (vector) embedding of node $v\in V_i$.
Then, the graph database can be seen as a set of sets, of the form $\mG = \{\mS_i\}_{i=1}^N$.

In this work, we aim to detect graph-level anomalies of two different types: (See Figure \ref{fig:background} for an illustration.)
\cbit
\item \textbf{Point graph anomaly}: defined as a graph that is a \textit{set containing anomalous nodes}, and
\item \textbf{Distribution graph anomaly}: defined as a graph that is an \textit{anomalous set} of not-necessarily-anomalous nodes.
\ceit 	

Correspondingly, we derive two different graph representations: the typical Mean-pooling and the newly-proposed MMD-pooling, described as follows. 

\subsubsection{\bf Mean-pooling for Point Anomalies}
\label{sssec:mean}
To detect graphs containing anomalous nodes, we simply use \text{\sf MEAN} as the \text{\sf READOUT} aggregation function in Eq. \eqref{eq:graphrep}, that is,

\vspace{-0.1in}
\beq
\label{eq:point}
\scalemath{0.9}{
\bh_{G_i} = \frac{1}{n_i} \sum_{l=1}^{n_i} \bh_{l,i}\;, \;\; \text{for all } G_i \in \mG \;.
}
\eeq
\vspace{-0.1in}

The intuition is that anomalous nodes would not only have significantly different embeddings, but also affect the embeddings of other nodes in their local neighborhood due to message-passing.
Mean-pooling would be effective in capturing sets with such anomalies as average is sensitive to outliers. 

\subsubsection{\bf MMD-pooling for Distribution Anomalies}
\label{sssec:mmd}

Differently, a distribution anomaly can arise from non-anomalous nodes, rendering mean-pooling ineffective.
Solely node-level detection approaches \cite{conf/sdm/DingLBL19,ocgnn20} would also fall short for the same reason.
Distinctly, \method 
takes into account the distribution information provided by each $\mS_i$ to identify such anomalies.

Suppose the samples (i.e. node embeddings) in each $\mS_i$ 
are distributed according to a (unknown) probability distribution $\bpi \in \mP$, where $\mP$ is the set of all probability distributions on the node embedding space.
Given two graphs $G_i$ and $G_{j}$, we define their similarity 
by a distribution kernel $\kappa(\cdot,\cdot)$ on $\mP$, i.e. $\kappa: \mP \times \mP \mapsto \R$, applied to their probability distributions as
\beq
\label{eq:kappa}
\kappa(\bpi,\bpip) = \langle \mu_{\bpi}, \mu_{\bpip}  \rangle_{\mH}\;,
\eeq
which is equal to the inner product between the kernel embeddings of $\bpi$ and $\bpip$ in  RKHS $\mH$, as given in Eq. \eqref{eq:meanembedeq}.

Based on Eq. \eqref{eq:mmdsq}, we can write the squared $\MMD$ as  
\beq
\scalemath{0.9}{
\MMD^2(\bp,\bq) 
= \langle \mu_{\bpi}, \mu_{\bpi}  \rangle_{\mH} + \langle \mu_{\bpip}, \mu_{\bpip}  \rangle_{\mH} - 2\langle \mu_{\bpi}, \mu_{\bpip}  \rangle_{\mH}
}
\eeq
and we can derive $\kappa(\bpi,\bpip)$ as equal to
\beq
\label{eq:distnkernel}
\hspace{-0.125in}
\scalemath{0.9}{
\langle \mu_{\bpi}, \mu_{\bpip}  \rangle_{\mH} = \E_{\bh,\bhp}[k(\bh,\bh')] = \int\int k(\bh,\bhp) d\bpi(\bh)d\bpip(\bhp).
}
\eeq
Given the (finite) sample sets $\mS_i$ and $\mS_{j}$, we can estimate the distribution kernel similarity in Eq. \eqref{eq:distnkernel} empirically, as
\beq
\label{eq:empkappa}
\scalemath{0.9}{
\kappa(\hbpi,\hbpip) = \frac{1}{n_i \cdot n_{j}} \sum_{u=1}^{n_i} \sum_{v=1}^{n_j} 
k(\bh_{u,i}, \bh_{v,j})
}
\eeq
where we use the (characteristic) Gaussian kernel for $k(\cdot,\cdot)$.

{We refer to Figure \ref{fig:background} for an intuitive comparison of MMD- vs. Mean-pooling, and an understanding of their complementary strengths. (See caption and Sec. \ref{sssec:meanvsmmd} for discussion.) 
}

{\bf Efficient and Explicit Graph Embedding:~} The kernel
 $\kappa(\cdot,\cdot)$ is a positive definite kernel on $\mP$. Given a graph database $\mG = \{\mS_i = \{\bh_{1,i}, \ldots, \;\bh_{n_i, i}\}\}_{i=1}^N$,
 we can use Eq. \eqref{eq:empkappa} to obtain the $(N\times N)$ empirical kernel (a.k.a. Gram) matrix $\bK$
 that can be directly input to the quadratic program (QP) solver for the dual OCSVM.
However, this would be expensive for graph databases with very large $N$, and even infeasible if $N$ is too large for $\bK$ to fit in  memory.

Apart from computational reasons, the kernel embedding of a distribution $\bp \in \mP$ as given in Def.n \ref{eq:meanembed}
is a {\em function} --rather than a \textit{vector} embedding-- in the RKHS of functions, just like a distribution in essence is a probability density function in the input space.
Having an explicit vector representation for each graph would provide flexibility, enabling the use of  other detectors.
 
therefore, we aim to obtain a decomposition of the Gram matrix $\bK = \bH \bH^T$ such that $\bH \in \R^{N\times r}$ can act as an empirical kernel map, where $r$ is the rank of $\bK$.
Then,
\beq
\bK_{ij} = \kappa(\hbpi,\hbpip) = \langle  \bH_i, \bH_j  \rangle
\eeq
which can be seen akin to $\langle \phi(G_i), \phi(G_j)\rangle$, that is the empirical inner product of the implicit embeddings in the RKHS as induced by the kernel, 
corresponding to Eq. \eqref{eq:kappa}. 
Notably, $\bH$ would consist of a vector embedding for each graph, i.e. $\bH_i \equiv h_{G_i}$, this time capturing characteristics w.r.t. the distribution of node embeddings.
Following earlier terminology, we refer to this procedure as \textit{MMD-pooling}.

The question is how to obtain such a decomposition. Eigendecomposition of $\bK = \bU \bS \bU^T$ is an option, where the empirical kernel map can be written as 
$\bH = \bU \bS^{1/2}$, i.e. as the eigenvectors of $\bK$ scaled by the square-root of their corresponding eigenvalues. However, it takes $O(N^2r)$ time  and $O(N^2)$ space as $\bK$ needs to be explicitly constructed in memory.
Therefore, it does not address the aforementioned efficiency challenges.

Instead, we use the \ny method, widely used for scaling kernelized algorithms 
\cite{conf/nips/WilliamsS00}.
Given 
$\mG$, it selects a subsample $\mB \subset \mG$ of size $k<N$,\footnote{\ny method's performance depends on the sampling scheme; we use random sampling for efficiency, with \textit{multiple} sample sizes as hyperparameter.} and provides a rank-$k$ approximation: 
\beq
\label{eq:ny}
{\bK} \;\;\approx\;\; \bK_{\mG,\mB} \; \bK_{\mB,\mB}^{-1} \;\bK_{\mG,\mB}^T 
\eeq

\noindent
where $\bK_{\mG,\mB} \in \R^{N\times k}$ is the Gram matrix between all examples in $\mG$ and those in the subsample $\mB$. Similarly, $\bK_{\mB,\mB} \in \R^{k\times k}$ is the Gram matrix between pairs of examples in $\mB$.
Note that the \ny method does not require the $(N\times N)$ $\bK$ matrix explicitly in memory.
It only necessitates the kernel computation between $Nk+k^2$ graph pairs, effectively down-scaling complexity to linear in the database size.
The matrix inverse $\bK_{\mB,\mB}^{-1}$ can also be carried out efficiently, considering $k$ is a small constant.
Upon eigendecomposition of the $(k\times k)$ matrix $\bK_{\mB,\mB} = \bV \bD \bV^T$, 
we can derive the approximation for the empirical kernel map $\bH$ as
\beq
\label{eq:empH}
\bH \;\;\approx\;\; \bK_{\mG,\mB} \; \bV \; \bD^{-1/2} \;.
\eeq
\hide{
Estimating $\bH$ is carried out by computing $\bK_{\mG,\mB}$ (from which $\bK_{\mB,\mB}$ can also be read out) via Eq. \eqref{eq:empkappa} in $O(Nk\bar{n}^2d')$  (where $\bar{n}$ depicts the average number of nodes in a graph, and $d'$ is the size of the node embeddings),\footnote{Although MMD-pooling is quadratic w.r.t. $\bar{n}$ (while mean-pooling is linear), it can be sped up via the subsample-based estimation of Eq. \eqref{eq:empkappa}.}
followed by the eigendecomposition of $\bK_{\mB,\mB}$ in $O(k^3)$, and finally a series of matrix products in Eq. \eqref{eq:empH} in $O(Nk^2)$.
The memory requirement can be written as $O(N(k+\bar{n}d'))$. 
}
This way MMD-pooling produces explicit graph-level embeddings efficiently.



\subsection{Anomaly Detection}
\label{ssec:anomaly}


Having obtained explicit vector embeddings of graphs, we train a one-class classifier, 
optimizing the Deep-SVDD objective \cite{RuffGDSVBMK18}: 
\begin{align}
\scalemath{1}{
\label{eq:obj}
    \min_\mW \quad \frac{1}{N} \sum_{i=1}^N \Vert \text{GIN}(G_i; \mW)  - \mathbf{c} \Vert_2^2 + \frac{\lambda}{2} \sum_{l=1}^L \Vert \bW^{(l)}\Vert^2_F
}
\end{align}
where $\text{GIN}(G_i)=\bh_{G_i}$ denotes the vector embedding for $G_i$
(based on Mean- or MMD-pooling),
$\bW^l$ denotes the (MLP) parameters of GIN at the $l$-th layer, $\mW=\{\bW^{(1)},\ldots,\bW^{(L)}\}$, $\mathbf{c}$ is the center of the hypersphere in the representation space (set to the average of all graph representations upon initializing the GIN), and finally $\lambda$ is the weight-decay hyperparameter.


As discussed in \cite{RuffGDSVBMK18}, deep SVDD classification suffers from ``hypersphere collapse'', where the trained model maps all input instances directly to the fixed center $\mathbf{c}$. We employ the regularizations proposed therein (no bias terms, etc.) to prevent this problem.

After training the model on all graphs, the distance to center is used as the anomaly score for each graph, that is
\beq
\label{eq:score}
score(G_i) =  \Vert \text{GIN}(G_i; \mW)  - \mathbf{c} \Vert_2\;.
\eeq

\textbf{Training and Hyperparameters:~}
Overall, \method consists of an $L$-layer GIN architecture (with 2-layer MLPs, see Eq. \eqref{eq:noderepgin}) for node embedding, followed by readout (i.e. Mean- or MMD-pooling), trained end-to-end via stochastic gradient descent, optimizing the deep SVDD objective at the output layer.

Apart from 
$\mG$, \method takes as input the values for a list of hyperparameters (HPs):
number of layers $L$, weight decay $\lambda$, learning rate $\eta$,  model seed (for parameter initialization).
Additionally it has 
\ny subsample size $k$ 
for MMD-pooling.
{For each HP, we define a grid of values (see Table \ref{table:hyperparams} in Appx.). We train \method separately using Mean- or MMD-pooling for graph embedding, using various HP configurations, each inducing a separate pool of candidate models.
The model pools are combined and then fed to our model selection module, which we describe next.}
A step-by-step outline of \method can be found in 
Algorithm \ref{alg:glam} in Appx. \ref{ssec:algo}.

\subsection{Unsupervised Model Selection}
\label{sec:supgpca}

\subsubsection{\bf Motivation}
Many popular anomaly detection models have various hyperparameters (HPs) to set (e.g. number of nearest neighbors $k$ for LOF \cite{breunig2000lof}, kernel bandwidth $\gamma$ and max. outlier fraction $\nu$ for OCSVM \cite{ScholkopfPSSW01}, etc.), which influence their performance \cite{Campos2016}. On the other hand, HP tuning (i.e. model selection) is notoriously hard for unsupervised anomaly detection in the absence of any labels.
The challenge is exacerbated for deep neural network (NN) based detectors, like \method, which automatically inherit a sizable list of HPs from NN models.
With the advent of deep learning, 
we believe this will be a growing challenge, yet
surprisingly, recent work 
on deep outlier models fall short in acknowledging and systematically addressing this challenge \cite{chalapathy2019learning,conf/sdm/DingLBL19}.
Some use hold-out (labeled) validation data for model tuning \cite{journals/corr/abs-2009-11732,ocgnn20}, 
which is not applicable for unsupervised settings.

In this work, we recognize that the inability to tune HPs effectively can get in the way of deep NN based detectors reaching their full potential. To that end, we carefully design and employ an unsupervised model selection (UMS) strategy for \method. 
To our knowledge, we are first to acknowledge UMS as a key challenge for {\em deep anomaly detection} and  build UMS into our proposed method.

\subsubsection{\bf Selection}
\label{ssec:selection}

Recently, two model selection techniques have been proposed for deep unsupervised disentangled representation learning, namely \udr \cite{DuanMSWBLH20} and ModelCentrality (MC) \cite{lin2020infogan}.\footnote{We found a couple of existing work on UMS for anomaly detection \cite{MarquesCZS15,journals/corr/Goix16} to be computationally too expensive and relatively much less effective.}
Both are simple consensus-based approaches, leveraging the agreement between models in the candidate pool to assign a ``reliability'' score to each model. 
We extend on these ideas for \method by fine-tuning the reliability scores recursively. We 
compare to \udr and MC in the experiments.


%

Specifically, 
we extend the idea of \mc \cite{lin2020infogan} by computing centrality \textit{recursively} based on a weighted bipartite network between candidate models and input graphs---wherein a  model gains higher centrality (i.e. reliability) the more they point with high anomaly score (edge weight) to graphs that are pointed by other high-centrality (reliable) models.

One of the earliest methods for computing centrality, namely hubness $h$ and authority $a$, of pages on the Web is the HITS algorithm \cite{kleinberg1999hits}.
Here, we employ this idea to estimate model ``reliability''
by constructing a complete bipartite network between $M$ candidate models and $N$ graphs, where

\vspace{0.005in}
\hspace{0.15cm}	$h_i \; \propto \; \text{sum of }  a_j\text{'s of all graphs } j \text{ that model } i \text{ points to} \;,$ 
	
\hspace{0.15cm}		$a_j \; \propto \; \text{sum of }  h_i\text{'s of all models } i \text{ that point to graph } j \;,$
which are estimated alternatingly over iterations.  
Upon convergence, the model with the largest hubness 
can be selected.
However, we recognize that 
this approach also provides an ensemble ranking of the graphs based on the final authority scores. 
We employ this strategy (called \hitse) for \method. 


%

\section{Experiments}
\label{sec:experiments}

\subsection{Setup}
\label{ssec:setup}
\subsubsection{\bf Datasets}
We evaluate \method on 15 public benchmark graph databases, 11 with node-labeled graphs and 4 containing node-attributed graphs.\footnote{\label{note1} All datasets are from TU Datasets: \url{https://chrsmrrs.github.io/datasets/docs/datasets/}. Our down-sampled versions can be found at \url{https://github.com/sawlani/GLAM}.} A summary of the datasets is given in Table \ref{table:data}.
Detailed descriptions can be found in Appx. \ref{ssec:datasets}. 
 
Our datasets are repurposed from binary graph classification datasets, where we designate one class as the inlier class, and  down-sample the other class(es) at $\approx$5\% to constitute the anomalous class.\footnoteref{note1}
For training and evaluation, we split each dataset into two; training data consists exclusively of inliers and test data contains both inliers and anomalies. 
Note that there exists {\bf no} validation set containing labeled anomalies, since we consider unsupervised detection.


\subsubsection{\bf Baselines} 
We compare to two types of baselines
\footnote{Baselines are part of a larger graph-level anomaly detection (GLAD) library, called PyGLAD, that we are building, to be released upon project completion.}:


\textit{{i)} Two-stage baselines} 
first use unsupervised graph-level embedding/kernel techniques to obtain vector/kernel representations of the graphs (stage 1: $\sbullet$ Weisfeiler-Lehman (\wl) kernel \cite{shervashidze2011weisfeiler}, $\sbullet$ Propagation kernel (\pk) \cite{neumann2016propagation}, and graph2vec (\gtovec) \cite{NarayananCVCLJ17}), and
 then employ point outlier detectors in the embedding/kernel space (stage 2: $\sbullet$ density-based LOF \cite{breunig2000lof} and $\sbullet$ one-class based OCSVM \cite{ScholkopfPSSW01}).
Note that 
\wl and  \gtovec apply to labeled graphs only.



\textit{ii) Node-level baselines:~} These are recent GNN based node anomaly detection methods in a {\em single} graph. We repurpose them to graph-level detection by scoring each graph with the average anomaly scores of its nodes.
\cbit
\item  OCGNN \cite{ocgnn20} builds on GNN-based node embeddings learned through the OCSVM objective; 
\item  DOMINANT \cite{conf/sdm/DingLBL19} employs a reconstruction-based loss for both graph structure and node attribute vectors.
\ceit

\begin{table}[!t]
\caption{Datasets in experiments.
Number of attributes in parentheses.
} 
\vspace{-0.1in}
\centering
\footnotesize{
\begin{tabular}{l|c|c|r|r}
	\toprule
	\rule{0pt}{2.5ex}\textbf{Name} & \textbf{Avg \#nodes} & \textbf{Type}  & $|$\textbf{Train}$|$ & $|$\textbf{Test}$|$ \\ \midrule
	\rule{0pt}{2.5ex}\textsc{Mixhop}  &         100          &    Node-Labeled     &         532          &         493         \\
	\textsc{Proteins}                 &        39.06         &    Node-Labeled     &         358          &         333         \\
	\textsc{Tox21}                    &        18.09         &    Node-Labeled     &         472          &         503         \\
	\textsc{Collab}                   &        74.49         &    Node-Labeled     &         395          &         397         \\
	\textsc{IMDB}                     &        19.77         &    Node-Labeled     &         270          &         240         \\
	\textsc{NCI1}                     &        29.87         &    Node-Labeled     &         1014         &        1096         \\
	\textsc{Mutagen}                  &        30.32         &    Node-Labeled     &         1189         &        1274         \\
	\textsc{Reddit}                   &        23.93         &    Node-Labeled     &         2418         &        2594         \\
	\textsc{DD}                       &        284.32        &    Node-Labeled     &         375          &         336         \\
	\textsc{AIDS-L}                   &        15.69         &    Node-Labeled     &         202          &         206         \\
	\textsc{DHFR-L}                   &        42.43         &    Node-Labeled     &         154          &         147         \\ \hline
	\rule{0pt}{2.5ex}\textsc{BZR}     &        35.75         & Attributed (3) &         170          &         154         \\
	\textsc{COX2}                     &        41.22         & Attributed (3) &         195          &         176         \\
	\textsc{AIDS-A}                   &        15.69         & Attributed (4) &         202          &         206         \\
	\textsc{DHFR-A}                   &        42.43         & Attributed (3) &         154          &         147         \\ \bottomrule
\end{tabular}
}
\vspace{-0.2in}
	\label{table:data} 
\end{table}

\noindent
\subsubsection{\bf Model Configurations}
Unsupervised model selection for anomaly detection has not drawn the necessary attention, and owing to the fact that most detectors rely on one or more hyperparameters (HPs), setting HP values is far from trivial.
To this end, we specify a grid of HPs for each of the baselines
(as listed in Appx. \ref{ssec:config}, Table \ref{table:hyperparams}),
and report the performance \textit{averaged} across all configurations.
This corresponds to the \textit{expected} performance when a configuration is selected at random from the grid (when no selection strategy is available).
We specify a HP-grid similarly for \method, which yields a pool of candidate models with various configurations.
We also train with both aggregation methods -- MMD and Mean -- and finally merge and feed both the corresponding pools to UMS.


\subsection{Results}
\label{ssec:results}

We conduct experiments to answer the following questions.
\bit
\item  {\em 1)} {\bf Detection performance:~} How effective is \method at graph-level anomaly detection, as compared to the two-stage and node-level baselines repurposed for the task? 
\eit
Two key novel components of \method are MMD-pooling and model selection. We also perform ablation studies to analyze their effect on detection performance.
\bit
\item {\em 2)} {\bf Mean- vs. MMD-pooling:~} Does MMD-pooling complement the traditional Mean-pooling? Can we leverage both of their strengths?
\item  {\em 3)} {\bf Model selection:~} Is unsupervised model selection effective as compared to expected (i.e. average) performance? Which selection methods are competitive? 
\eit

\subsubsection{\bf Detection Performance}

\begin{table*}[!ht]
\caption{Anomaly detection performance of all methods.
For baselines, we run the experiment over a grid of hyperparameters$^5$ 
and report the average and stand. dev. of ROC-AUC scores. As \method employs (unsupervised) model selection, it only outputs \underline{\bf one} ranking, and we simply report its ROC-AUC.  Per dataset rank provided in parentheses (the lower the better). Average performance and 
rank across datasets given in the last rows. Symbols $\blacktriangle$ and $\triangle$ denote the cases where \method is significantly better than
baseline w.r.t. the Wilcoxon signed rank test, p$<$0.01 and p$<$0.1 respectively. (O.O.M.: out-of-memory)
} 
\centering
	\vspace{-0.1in}	
	\scalebox{0.84}{
\begin{tabular}{l|c|c|c|c|c|c|c|c||c}
	\toprule
	\multicolumn{1}{c|}{Dataset}     &         \pk-LOF          &        \pk-OCSVM         &         \wl-OCSVM          &        \wl-LOF         &       \gtovec-LOF        &      \gtovec-OCSVM       &         DOMINANT         &          OCGNN           &               GLAM                \\ \midrule
	\rule{0pt}{2.5ex}\textsc{Mixhop} &     0.67 $\pm$ 0.17(5)      &     0.69 $\pm$ 0.14(4)      &     0.72 $\pm$ 0.09(3)    &     0.57 $\pm$ 0.01(7)      &     0.67 $\pm$ 0.08(5)      &     0.51 $\pm$ 0.03(8)      & \textbf{1.00 $\pm$ 0.00(1)} &            N/A             &               0.99(2)               \\
	\textsc{Proteins}                &     0.68 $\pm$ 0.12(5)      &     0.67 $\pm$ 0.09(6)      &     0.72 $\pm$ 0.05(4)      &     0.78 $\pm$ 0.03(2)      &     0.47 $\pm$ 0.05(8)      &     0.46 $\pm$ 0.09(9)      &     0.48 $\pm$ 0.23(7)      &     0.75 $\pm$ 0.34(3)      &           \textbf{0.82(1)}          \\
	\textsc{Tox21}                   &     0.53 $\pm$ 0.05(5)      &     0.57 $\pm$ 0.04(4)      &     0.65 $\pm$ 0.05(3)      &     0.66 $\pm$ 0.02(2)      &     0.47 $\pm$ 0.06(8)      &     0.47 $\pm$ 0.06(8)      &     0.53 $\pm$ 0.02(5)      &     0.52 $\pm$ 0.05(7)      &           \textbf{0.70(1)}           \\
	\textsc{Collab}                  &     0.73 $\pm$ 0.05(7)      &     0.78 $\pm$ 0.02(5)      &     0.81 $\pm$ 0.06(4)      &     0.84 $\pm$ 0.00(3)      &     0.54 $\pm$ 0.18(9)      &     0.75 $\pm$ 0.03(6)      &     0.85 $\pm$ 0.02(2)      &     0.58 $\pm$ 0.22(8)      &           \textbf{0.87(1)}           \\
	\textsc{IMDB}                    &     0.61 $\pm$ 0.09(6)      &     0.68 $\pm$ 0.06(2)      &     0.65 $\pm$ 0.04(3)      &     0.63 $\pm$ 0.03(4)      &     0.45 $\pm$ 0.09(8)      &     0.35 $\pm$ 0.04(9)      &     0.62 $\pm$ 0.01(5)      &     0.51 $\pm$ 0.05(7)      &           \textbf{0.70(1)}           \\
	\textsc{NCI1}                    &     0.64 $\pm$ 0.06(2)      &     0.45 $\pm$ 0.07(7)      & \textbf{0.73 $\pm$ 0.04(1)} &     0.60 $\pm$ 0.08(3)      &     0.48 $\pm$ 0.10(6)      &     0.33 $\pm$ 0.11(9)      &     0.49 $\pm$ 0.12(5)      &     0.45 $\pm$ 0.16(7)      &               0.59(4)                \\
	\textsc{Mutagen}                 &     0.57 $\pm$ 0.05(4)      &     0.53 $\pm$ 0.02(7)      &     0.56 $\pm$ 0.03(6)      &     0.52 $\pm$ 0.03(8)      & \textbf{0.64 $\pm$ 0.04(1)} &     0.57 $\pm$ 0.05(4)      &     0.59 $\pm$ 0.07(2)      &     0.49 $\pm$ 0.06(9)      &               0.59(2)                \\
	\textsc{Reddit}                  &     0.45 $\pm$ 0.06(6)      &     0.75 $\pm$ 0.02(3)      &     0.54 $\pm$ 0.14(5)      & \textbf{0.83 $\pm$ 0.01(1)} &     0.43 $\pm$ 0.07(7)      &     0.58 $\pm$ 0.04(4)      &            O.O.M.             &            O.O.M.             &               0.76(2)                \\
	\textsc{DD}                      &     0.83 $\pm$ 0.06(4)      &     0.76 $\pm$ 0.04(5)      & \textbf{0.92 $\pm$ 0.01(1)} &     0.91 $\pm$ 0.01(2)      &     0.31 $\pm$ 0.03(8)      &     0.46 $\pm$ 0.03(7)      &            O.O.M.             &     0.60 $\pm$ 0.06(6)      &               0.84(3)                \\
	\textsc{AIDS-L}                  &     0.77 $\pm$ 0.11(4)      &     0.73 $\pm$ 0.10(5)      & \textbf{0.95 $\pm$ 0.04(1)} &     0.94 $\pm$ 0.07(2)      &     0.30 $\pm$ 0.16(8)      &     0.12 $\pm$ 0.12(9)      &     0.68 $\pm$ 0.04(6)      &     0.42 $\pm$ 0.14(7)      &               0.91(3)                \\
	\textsc{DHFR-L}                  &     0.64 $\pm$ 0.07(5)      &     0.74 $\pm$ 0.05(4)      &     0.77 $\pm$ 0.07(3)      &     0.81 $\pm$ 0.05(2)      &     0.62 $\pm$ 0.12(6)      &     0.40 $\pm$ 0.08(7)      &     0.36 $\pm$ 0.07(9)      &     0.37 $\pm$ 0.14(8)      &           \textbf{0.83(1)}           \\
	\textsc{BZR}                     & \textbf{0.69 $\pm$ 0.11(1)} &     0.58 $\pm$ 0.06(2)      &            -             &            -             &            -             &            -             &     0.37 $\pm$ 0.01(5)      &     0.55 $\pm$ 0.07(3)      &               0.55(3)                \\
	\textsc{COX2}                    &     0.70 $\pm$ 0.05(2)      &     0.55 $\pm$ 0.10(4)      &            -             &            -             &            -             &            -             &     0.42 $\pm$ 0.02(5)      &     0.56 $\pm$ 0.13(3)      &           \textbf{0.71(1)}           \\
	\textsc{AIDS-A}                  &     0.53 $\pm$ 0.12(3)      &     0.84 $\pm$ 0.04(2)      &            -             &            -             &            -             &            -             &     0.05 $\pm$ 0.00(5)      &     0.43 $\pm$ 0.11(4)      &           \textbf{0.94(1)}           \\
	\textsc{DHFR-A}                  &     0.70 $\pm$ 0.06(2)      &     0.62 $\pm$ 0.03(3)      &            -             &            -             &            -             &            -             &     0.43 $\pm$ 0.00(5)      &     0.58 $\pm$ 0.06(4)      &           \textbf{0.82(1)}           \\ \midrule
	\textit{avg (labeled)}           & \textit{0.65 $\pm$ 0.11} & \textit{0.67 $\pm$ 0.11} & \textit{0.73 $\pm$ 0.13} & \textit{0.74 $\pm$ 0.14} & \textit{0.49 $\pm$ 0.13} & \textit{0.47 $\pm$ 0.15} & \textit{0.62 $\pm$ 0.20} & \textit{0.52 $\pm$ 0.11} & \textit{\textbf{0.78 $\pm$ 0.13}} \\
	\textit{avg (all)}               & \textit{0.65 $\pm$ 0.10} & \textit{0.66 $\pm$ 0.11} &            -             &            -             &            -             &            -             & \textit{0.53 $\pm$ 0.24} & \textit{0.52 $\pm$ 0.10} & \textit{\textbf{0.77 $\pm$ 0.13}} \\
 	\textbf{avg rank} & $\textbf{4.07}^\blacktriangle$  &
 	\textbf{4.20}$^\blacktriangle$ & \textbf{3.09} & \textbf{3.27}$^\triangle$ & \textbf{6.73}$^\blacktriangle$ & \textbf{7.27}$^\blacktriangle$ & \textbf{4.77}$^\blacktriangle$ & \textbf{5.85}$^\blacktriangle$ & \textbf{1.8}
	\\ \bottomrule
\end{tabular}
	}
	\vspace{-0.15in}
	\label{table:horserace} 
\end{table*}

Table~\ref{table:horserace} gives the ROC-AUC 
performances for \method and all the baselines on each dataset.\footnote{DOMINANT threw out-of-memory error on {\sc Reddit} and {\sc DD}. OCGNN threw that error on {\sc Reddit}, and is incompatible with \textsc{Mixhop}'s dense format.} 

On average across HPs, \method outperforms all the baselines. Owing to different characteristics of the datasets, we do not expect \method to be the best model on every dataset. However, we observe that this is still true on a majority (8 out of 15) of the datasets. Moreover, no baseline stands out as a clear runner-up -- some baselines which exhibit the best performance for a dataset can be seen to be arbitrarily bad on others.

\begin{figure}[!ht]
	\vspace{-0.1in}
	\centering
	\includegraphics[width=0.8\linewidth]{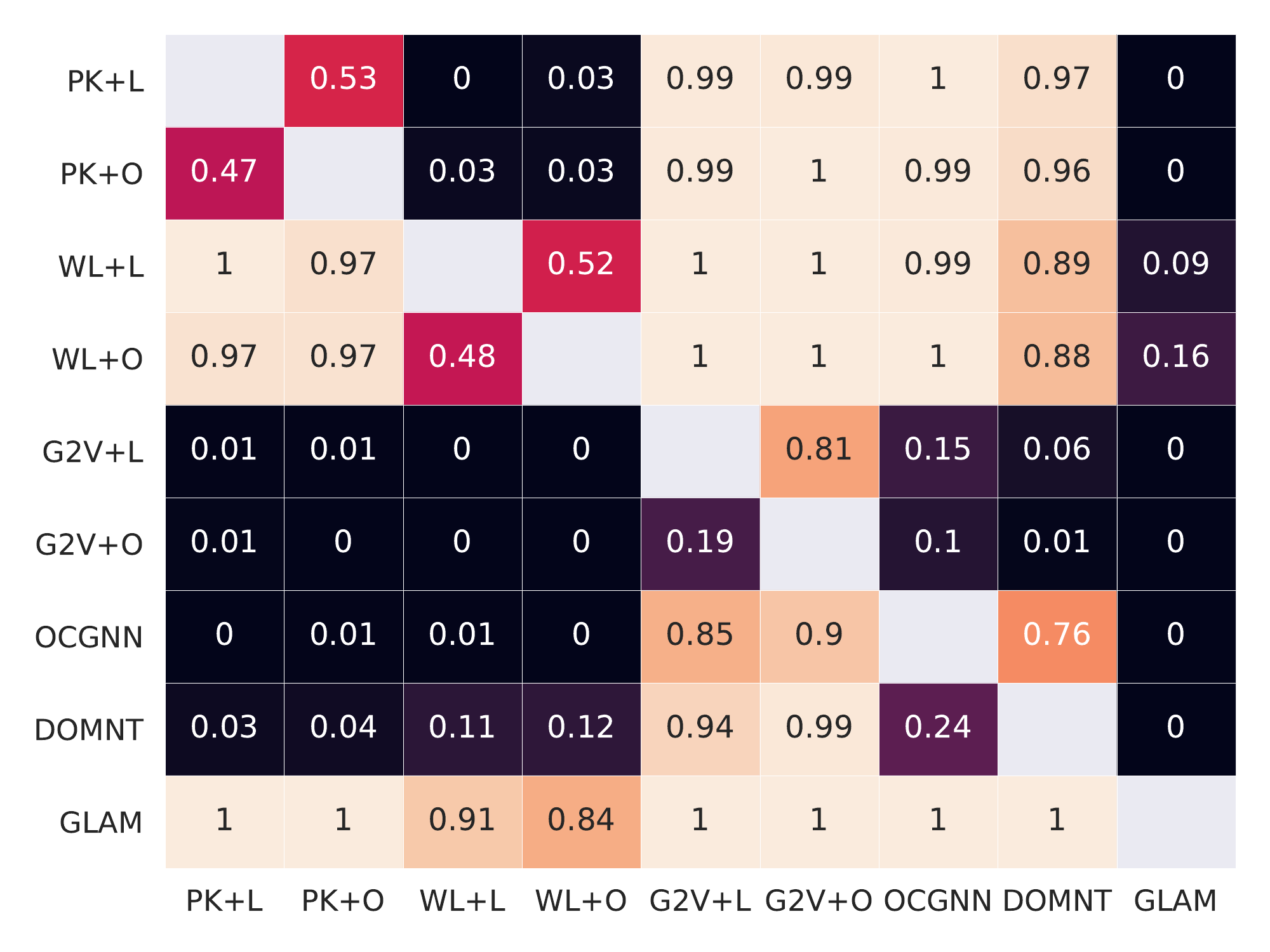}
	\vspace{-0.2in}
	\caption{Comparison of detection methods by one-sided paired Wilcoxon signed-rank test. $p$-values smaller than $0.05$ correspond to the cases where col-method is significantly better than the row-method.\label{fig:wilcoxon_horse_race}} 
	\vspace{-0.15in}
\end{figure}

The comparison can be further demonstrated in Figure~\ref{fig:wilcoxon_horse_race}, which displays the results of pairwise {one-sided} Wilcoxon signed rank tests between all competing methods.
The entry in cell $(i,j)$ is the $p$-value, given the (alternative) hypothesis that method $j$ performs better than method $i$ against the null (that they are not different).
Each paired test is done on the ROC-AUC values from 15 datasets, except tests involving  \wl and \gtovec are only over 11 (node-labeled) datasets, as those do not apply to attributed graphs.

Not only does \method outperform all the baselines on average (as shown in Table \ref{table:horserace}),
as these test results show, the difference is also significant at $p\leq 0.01$ against all baselines but \wl.
\wl appears more competitive than other baselines, however $p$-values are still fairly low (0.09 w/ LOF and 0.16 w/ OCSVM), and it comes with the caveat that it only applies to node-labeled graphs. 
Based on these results, we conclude that end-to-end graph-level detection with \method is more effective than these existing baselines.

\footnotetext{Detailed hyperparameters for each baseline are in Table \ref{table:hyperparams} in Appx. \ref{ssec:config}.}

\subsubsection{\bf Mean-pooling vs. MMD-pooling}
\label{sssec:meanvsmmd}

Viewing a graph as a distribution over node embeddings, MMD-pooling 
offers the capacity to detect graphs with an anomalous ``composition'', which is beyond the scope of Mean-pooling.
On the other hand, Mean-pooling would still be preferable when a graph contains only a few anomalous nodes---as this may not shift the distribution largely enough for MMD to detect.
Hence, the two techniques are \textit{complementary}, enabling both point and distribution anomaly detection.

\begin{table}[!t]
\vspace{-0.05in}
	\caption{(left) Mean$\pm$std. performance of candidate models (avg'ed over all HP configs) based on MMD- vs. Mean-pooling. (right) 
	Performance change (\%) after model selection from Mean-pool \textit{only} vs. MMD-pool \textit{only} (rathen than both). MMD- and Mean- pooling are complementary, boosting overall performance.} 
	\centering
	\vspace{-0.125in}
	\scalebox{0.85}{
		\hspace{-0.1in}
	\scalebox{0.84}{
	\begin{tabular}{l||c|c|}
		\toprule
		\multicolumn{1}{c||}{\textbf{Dataset}} &    \textbf{MMD-pool}     &    \textbf{Mean-pool}    \\ 
		\midrule
		\rule{0pt}{2.5ex}\textsc{Mixhop}       &     0.92 $\pm$ 0.07      & \textbf{0.98 $\pm$ 0.02}     \\
		\textsc{Proteins}                      & \textbf{0.79 $\pm$ 0.04} &     0.71 $\pm$ 0.11           \\
		\textsc{Tox21}                         &     0.60 $\pm$ 0.04      & \textbf{0.70 $\pm$ 0.04}       \\
		\textsc{Collab}                        &     0.84 $\pm$ 0.04      & \textbf{0.87 $\pm$ 0.07}      \\
		\textsc{IMDB}                          & \textbf{0.62 $\pm$ 0.06} &     0.59 $\pm$ 0.05           \\
		\textsc{NCI1}                          &     0.56 $\pm$ 0.04      & \textbf{0.62 $\pm$ 0.04}      \\
		\textsc{Mutagen}                       &     0.54 $\pm$ 0.06      & \textbf{0.61 $\pm$ 0.06}      \\
		\textsc{Reddit}                        & \textbf{0.70 $\pm$ 0.06} &     0.55 $\pm$ 0.07          \\
		\textsc{DD}                            &     0.70 $\pm$ 0.07      & \textbf{0.85 $\pm$ 0.04}      \\
		\textsc{AIDS-L}                        &     0.79 $\pm$ 0.07      & \textbf{0.86 $\pm$ 0.07}      \\
		\textsc{DHFR-L}                        &     0.77 $\pm$ 0.08      & \textbf{0.79 $\pm$ 0.06}     \\
		\textsc{BZR}                           &     0.50 $\pm$ 0.12      & \textbf{0.69 $\pm$ 0.05}     \\
		\textsc{COX2}                          &     0.65 $\pm$ 0.10      & \textbf{0.70 $\pm$ 0.07}      \\
		\textsc{AIDS-A}                        & \textbf{0.83 $\pm$ 0.10} &     0.64 $\pm$ 0.12            \\
		\textsc{DHFR-A}                        &     0.72 $\pm$ 0.09      & \textbf{0.84 $\pm$ 0.05}
		\\
		\bottomrule 
		\multicolumn{3}{c}{\vspace{0.07in}}

	\end{tabular}}
	\hspace{0.15in}\hfill
	\scalebox{0.84}{
	\begin{tabular}{rc|c|}
	\toprule
\method :  &	\textbf{w/o MMD}    &  \textbf{w/o Mean} \\ 
	\midrule
  &  1.00     &     0     \\
  &  -6.49    &    1.20      \\
  &  0        &     -1.11      \\
   &   9.09     &     -11.11      \\
  &    3.33     &     -1.16      \\
  &    -9.375   &     -4.47      \\
   &   6.35     &     -1.72      \\
   &   0        &     -1.22      \\
   &   10.60   &     -7.27      \\
  &    -40.74    &     5.00     \\
   &   5.61    &     -3.70      \\
   &   -42.42    &     0      \\
  &    21.43   &     16.67      \\
  &    1.39    &     4.05     \\
  &    5.75    &     1.20      \\
 \midrule
avg : & \textbf{-2.29}    &     \textbf{-0.24}     \\
 \bottomrule
\end{tabular}}
	}
	\vspace{-0.3in}
	\label{table:MMDvMean} 
\end{table}

In Table \ref{table:MMDvMean} (left), we compare the performance of the models in the MMD vs. Mean pools on average (avg.'ed across HPs) for all 15 datasets. 
The better pooling technique is not consistent across datasets. Moreover, the differences can be quite large in both directions, showcasing their complementary strengths.
To leverage both,
\method combines these two pools, over which model selection is performed.
As shown in Table \ref{table:MMDvMean} (right), using \textit{both} Mean- \textit{and} MMD-pooling achieves improved results as compared to vanilla Mean-only or MMD-only pooling. We see that there is a drop in performance when model selection excludes Mean-pooling and an even bigger drop in performance when it excludes MMD-pooling.

\textbf{Case Study:~}
Next we take a closer look at the anomalous graphs detected by Mean- vs. MMD-pooling. Figure \ref{fig:background} shows the 
node embedding spaces by MMD- vs. Mean-pooling, respectively, on \textsc{Proteins}. Inlier nodes (green) span the whole background, as expected, while those of MMD anomalous graphs (red, orange) cluster in small zones, exhibiting distinct distributions. MMD-anomalies also look quite distinct visually. Notice that Mean-pooling alone, without the distributional ``lens'', falls short in spotting them.

\hide{
\subsubsection{\bf Sensitivity to Hyperparameters} 


Two-stage approaches do not have many hyperparameters, which makes them relatively stable.
\saurabh{this argument is proving hard to make}

\saurabh{unparametrized ones especially - the argument that bad models don't look alike doesn't hold in such algorithms, hence model selection, although helpful, is less effective than in end-to-end parametrized methods.}

\saurabh{Figure X displays the distribution of ROC-AUC's for each of the baselines as well as \method.
From the figure, we can see that the performance of both two-stage and end-to-end models is sensitive to the chosen hyperparameters. In the case of parametrized models, this sensitivity extends to the the random seeds chosen during initialization.

This observed sensitivity to hyperparameters necessitates the need for model selection,
which can be a challenging task in itself in an unsupervised setting. We tackle this in the following section.}
}

\subsubsection{\bf Model Selection}

\begin{figure}[!ht]
	\vspace{-0.2in}
	\centering
	\includegraphics[width=0.75\linewidth]{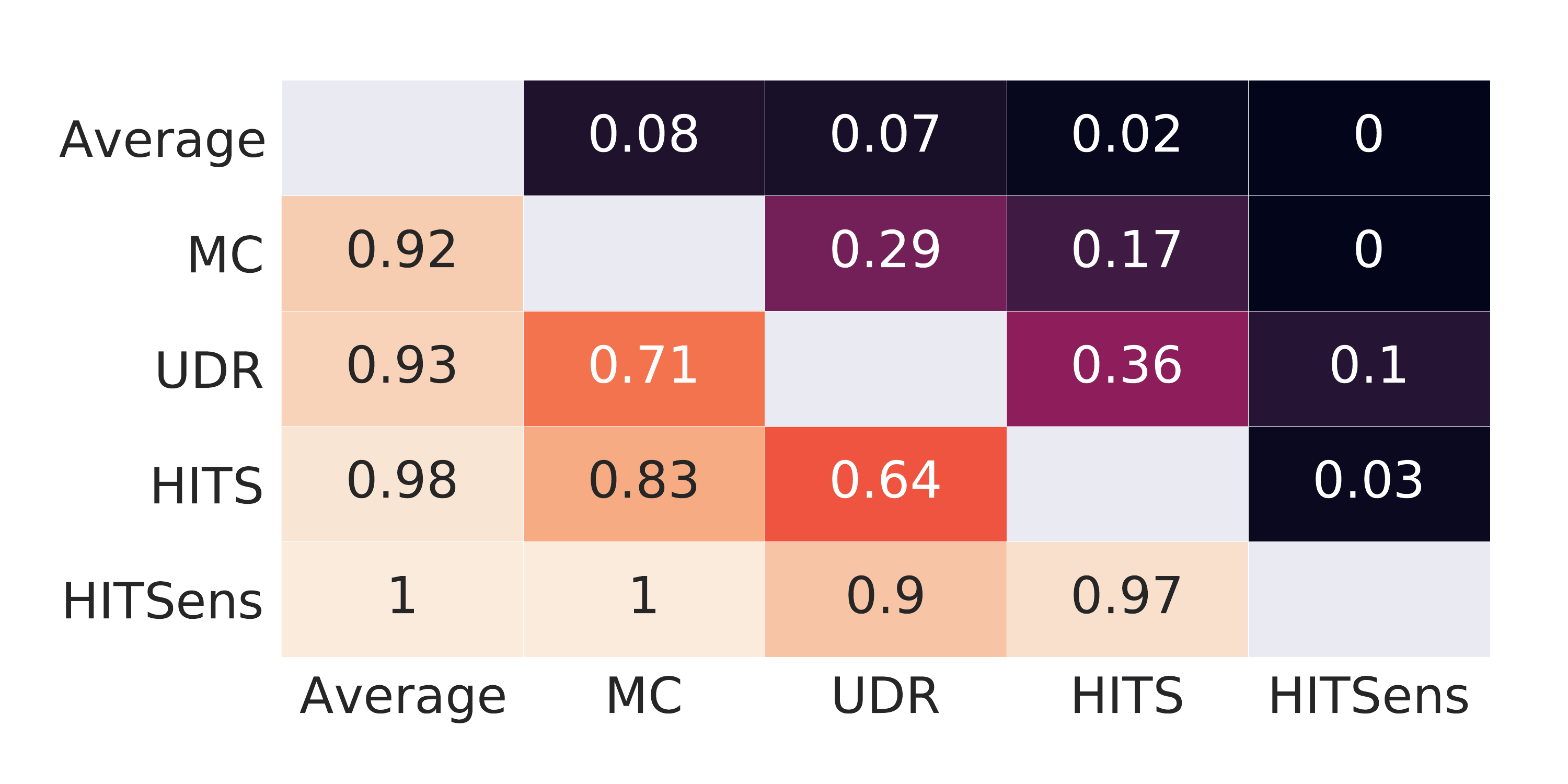}
	\vspace{-0.15in}
	\caption{Comparison of selection methods. Improvement over average performance is significant at $p\leq 0.1$. \label{fig:wilcoxon_selection}}
		\vspace{-0.075in}
\end{figure}

To analyze the benefits of employing unsupervised model selection techniques, in Figure \ref{fig:wilcoxon_selection} we compare (using Wilcoxon signed rank test, as before) the performance of the model as selected by \udr, \mc, \hits, or \hitse across datasets.
We also include the average model in this comparison.
We find that all strategies are competitive, significantly outperforming Average at $p\leq 0.1$.
\hitse, the consensus ranking by HITS that \method employs, outperforms others and is the most competitive.


\begin{figure}[!ht]
	\vspace{-0.1in}
	\centering
	\includegraphics[width=0.99\linewidth]{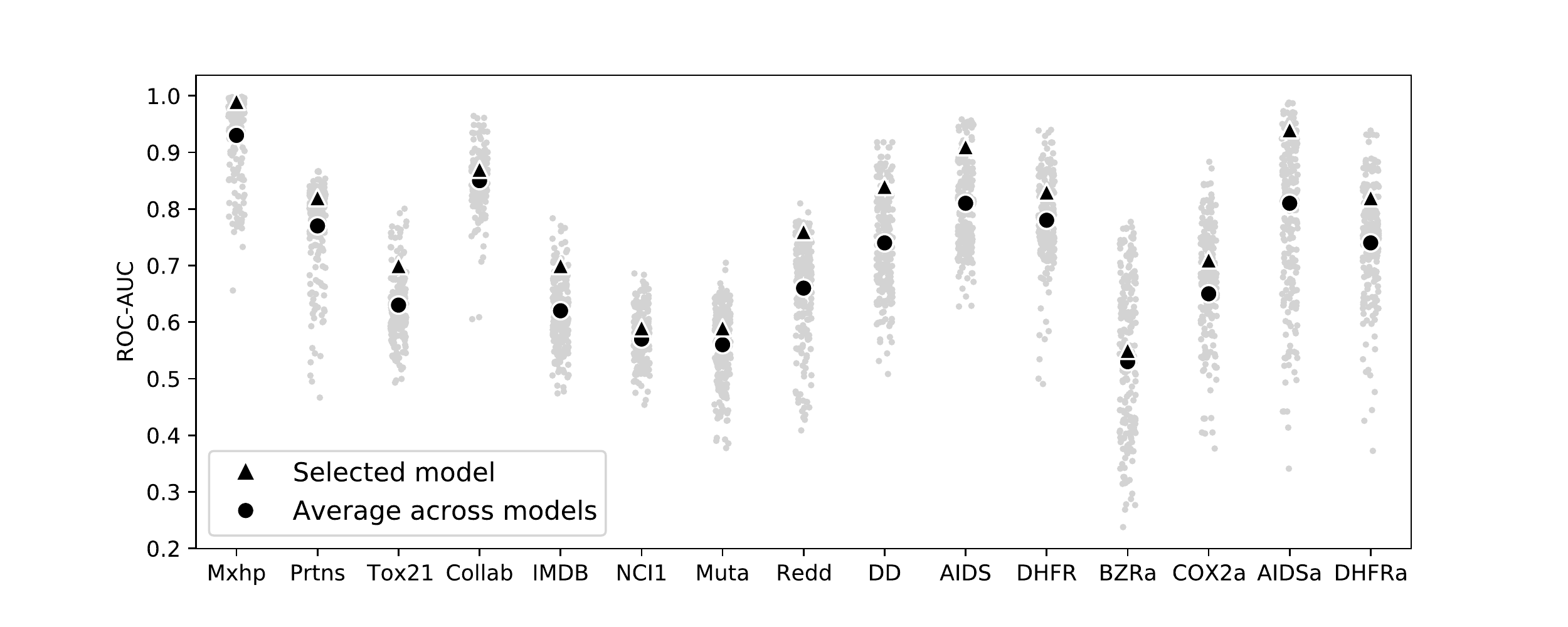}
	\vspace{-0.3in}
	\caption{Performance of  \method models w/ different HP configs in the pool. \method (w/ selection) consistently improves over Avg. (i.e. random choice). \label{fig:selection}}
		\vspace{-0.05in}
\end{figure}

In Figure \ref{fig:selection}, we show the performance for all candidate models in the \method pool (Mean+MMD) per dataset (gray dots), where Average (circle) and the \method-selected model's performance (triangle) are marked.
Notice that the pools consist of models with a large variation of performance, demonstrating the potential value for selection. Notably, \method (with selection) is consistently similar to or better than Average, providing up to 16\% improvement. 

\section{Related Work}
\label{sec:related}

%
%


\hide{
\method is designed for detecting anomalous graphs within a database, containing graphs of varying sizes, whose nodes are not necessarily aligned, and that may exhibit complex properties such as node labels (discrete), a set of node attributes (numeric), directed and/or weighted edges.
To our knowledge, there is \textit{no existing work on graph-level anomaly detection that can seamlessly handle graphs of such nature}. (See Table \ref{tab:related}.)
}

{\em Point-cloud outlier detection (OD).~} OD for 
vector data has a large literature with a vast array of approaches (distance, density, depth, angle, etc. based) \cite{aggarwal2015outlier}. With the advent of deep neural networks, 
deep outlier detection has drawn attention recently \cite{chalapathy2019learning,journals/corr/abs-2009-11732}.
These do not (at least directly) apply to  anomaly detection for graphs, which exhibit relational structure. 

\textit{Anomaly detection in a single graph.~} 
Majority of the literature on graph anomaly detection
focuses on node/edge/subgraph anomalies within a \textit{single} graph (see \cite{AkogluTK15} for survey).
Several recent techniques based on graph neural networks (GNNs) also fall into this category. 
Netwalk \cite{conf/kdd/YuCAZCW18}, DOMINANT \cite{conf/sdm/DingLBL19} and  OCGNN \cite{ocgnn20} aim to detect anomalies through \textit{node-level} representations.
Similarly, numerous representation learning techniques for graphs, such as DeepWalk \cite{deepwalk2014}, GraphSage \cite{conf/nips/HamiltonYL17}, more recently GAL \cite{conf/cikm/0003DYJW020}, and many others  \cite{hamilton2017representation} focus on \textit{node-level} embeddings.
How to aggregate these for graph-level anomaly detection is unclear.



\textit{Graph-level anomaly detection in a graph database.~}
The literature on \textit{graph-level} anomaly detection is relatively much sparser.
Only a few techniques exist for detecting anomalies within a set or series of graphs \cite{conf/kdd/NobleC03,conf/kdd/ManzoorMA16,conf/kdd/EswaranFGM18,zhao2021using}.
Most are traditional solutions that rely on substructure or motif counts, and unlike GNN-based approaches, cannot seamlessly handle graphs with various properties. For example, SpotLight \cite{conf/kdd/EswaranFGM18} cannot accommodate node attributes,  Subdue \cite{conf/kdd/NobleC03} and StreamSpot \cite{conf/kdd/ManzoorMA16} cannot handle weighted edges or multi-attributed nodes.
A recent work using deep learning \cite{zhao2021using} did not tackle distribution anomalies or  model selection and rather focused on evaluation issues with classification datasets.

%

\textit{Graph-level representation learning.~}
A body of 
 methods embeds an entire graph with complex properties, including Patchy-San \cite{pmlr-v48-niepert16}, graph2vec \cite{NarayananCVCLJ17}, and GIN \cite{conf/iclr/XuHLJ19}. 
In addition, various graph kernels \cite{shervashidze2011weisfeiler,yanardag2015deep,neumann2016propagation} either provide a vector representation for each graph or pairwise similarities between graphs 
 \cite{journals/corr/abs-1904-12218}.
However, these do not explicitly tackle anomaly detection. 
Their output (embeddings or distances)
 need to be input to a certain choice of an off-the-shelf outlier detector. 
In such a two-stage process, the first stage
(representation learning) is disconnected from the final goal (anomaly detection).

\hide{
In contrast, \method is designed for \textit{end-to-end} anomaly detection where the representation learning is inherently aligned with the objective of anomaly detection. 
Moreover, by relying on a GNN-based architecture, it can effectively handle graphs with a variety of different properties.
We are not aware of any existing GNN-based end-to-end solution to graph-level anomaly detection.
}

\textit{Unsupervised model selection for anomaly detection.~}
GNNs exhibit flexibility and expressiveness, however, these do not come for free---GNN-based models have many hyperparameters (HPs) that influence their performance \cite{journals/corr/abs-1811-05868}. 
Model tuning for (semi-)supervised tasks  can be handled based on hold-out labeled data, while there exists no such guidance for unsupervised tasks. 
As such, unsupervised model selection for GNN-based anomaly detection remains an open challenge.

Many detectors are sensitive to the choice of HP values,
\cite{Campos2016}, yet, there exists limited work on UMS for anomaly detection \cite{MarquesCZS15,journals/corr/Goix16}. These employ \textit{internal} model evaluation strategies using input and/or outlier scores only.
There are also a couple work on UMS for deep disentangled representation learning \cite{DuanMSWBLH20,lin2020infogan}, which we adapted and compared to 
for anomaly detection for the first time.
Despite the advent of deep detectors \cite{chalapathy2019learning,journals/corr/abs-2009-11732}, with numerous HPs, the subject of model selection has not raised the attention it demands.



\section{Conclusion}
\label{sec:conclusion}

  

  %
  
In this work we presented three main contributions.
First we proposed \method, a novel Graph-Level Anomaly detection Model based on GNNs.
\method employs an end-to-end 
 anomaly detection objective and can admit 
unordered graphs of varying sizes without node correspondence. 
Second, different from prior work on node/edge/subgraph anomalies, we targeted {distribution} graph anomalies  as graphs with anomalous sets of nodes. To this end, we introduced MMD-pooling 
toward capturing distributional patterns of node embeddings in a graph database.
Finally, we strived to systematically address the unsupervised model selection (UMS) problem, a key challenge especially for deep neural network based models (with a sizable list of hyperparameters) as well as unsupervised anomaly detection (without access to ground-truth labels) -- which may be of independent value for the anomaly mining community.  
Extensive experiments 
showed that 
\method significantly outperforms key baselines  in expectation across varying hyperparameter values. 
We also showed that both MMD-pooling and UMS are key players in \method's effectiveness.
To foster progress on these important fronts, we open-source all of our code and datasets  at \url{https://github.com/sawlani/GLAM}. 
\vspace{-1mm}

\section*{Acknowledgments}{
This work is sponsored by NSF CAREER 1452425. We also thank PwC Risk and Regulatory Services Innovation Center at Carnegie Mellon University. Any conclusions expressed in this material are those of the author and do not necessarily reflect the views, expressed or implied, of the funding parties.
}

\bibliographystyle{IEEEtran}

\bibliography{refs,ref}


\appendix 
\section{Appendix}
\label{sec:appendix}

\subsection{Step-by-Step Outline for \method}
\label{ssec:algo}
The pseudo-code outlining the individual steps of our proposed \method is given as in Algorithm \ref{alg:glam}.

\hspace{-.2in}
\begin{minipage}{\linewidth}

\begin{algorithm}[H] 
	\caption{\method: Graph-Level Anomaly detection Model}
	\label{alg:glam}
	\begin{algorithmic}[1]
		\small{
		\REQUIRE Graph database $\mG$ (containing $N$ plain/node-labeled/attributed graphs), Hyper-param.s: \#layers $L$,  weight-decay $\lambda$, learn.-rate $\eta$, model seed, \ny subsample size $k$ 
		\ENSURE Anomaly scores for all $G_i\in \mG$
		
		\STATE Specify a pool of configurations $\mC$ for hyper-param.s  (Table \ref{table:hyperparams})
		\STATE $Pool := \emptyset$, initialize the model pool
		\FOR{{\bf each} unique configuration $C\in \mC$} 
		\STATE Node embedding via GIN in Eq \eqref{eq:noderepgin}, 
		$\forall G_i\in \mG$
				\FOR[graph embedding]{{\bf each} pooling strategy}
							\STATE  $h_{G_i}^{\text{Mean}}$ via Eq \eqref{eq:point} (\S\ref{sssec:mean}), $\forall G_i\in \mG$ \COMMENT{\text{Mean}-pooling}
				 			\STATE  $h_{G_i}^{\text{MMD}}$ via Eq \eqref{eq:empH} (\S\ref{sssec:mmd}), $\forall G_i\in \mG$ \COMMENT{{MMD}-pooling}
				\ENDFOR
				\STATE End-to-end training $M^{\text{Mean},C}$ $:=$ \method($\{h_{G_i}^{\text{Mean}}\}_{i=1}^N$, config.s $C$) optimizing \eqref{eq:obj} (Sec. \ref{ssec:anomaly})
				\STATE Anomaly scores $S^{\text{Mean},C} \in \R^N$ using $M^{\text{Mean},C}(\mG)$,  Eq \eqref{eq:score}
				\STATE End-to-end training $M^{\text{MMD},C}$ $:=$ \method($\{h_{G_i}^{\text{MMD}}\}_{i=1}^N$, config.s $C$) optimizing \eqref{eq:obj}
				\STATE Anomaly scores $S^{\text{MMD},C} \in \R^N$ using $M^{\text{MMD},C}(\mG)$, Eq. \eqref{eq:score}
				\STATE $Pool := Pool \cup S^{\text{Mean},C} \cup S^{\text{MMD},C}$
		\ENDFOR
		\STATE Model selection w/ \hitse on the $Pool$ (\S\ref{ssec:selection})
		\STATE {\bf return} anomaly scores $S \in \R^N$ by \hitse 
	}
	\end{algorithmic}
\end{algorithm}
\end{minipage}

\vspace{0.05in}
\subsection{Dataset Description}
\label{ssec:datasets}
We use 15 datasets for evaluation. \textsc{Mixhop} 
graphs exhibit the Barabasi-Albert structure with 5 different node labels. A ``homophily'' parameter is specified for a graph, which dictates the probability with which a node connects to another node of the same label. In our case, inlier and outlier graphs have homophily 0.7 and 0.3, respectively. 
\textsc{AIDS}, \textsc{BZR}, \textsc{COX2}, \textsc{DHFR}, \textsc{Mutagenicity} and \textsc{Tox21\_AhR} are datasets that consist of small molecules where node labels indicate the specific atom. 
Each dataset consists of two classes, corrresponding to whether or not the molecule exhibits a certain property. All the above have inlier class as $0$, with the exception of \textsc{Tox21\_AhR}, whose inlier class is $1$.
Among these, \textsc{AIDS}, \textsc{BZR}, \textsc{COX2} and \textsc{DHFR} are node-attributed datasets.
\textsc{DD} and \textsc{PROTEINS} are bioinformatics classification datasets, also with classes indicating a physical property. $0$ is the inlier class in both the datasets.
\textsc{COLLAB}, \textsc{IMDB-BINARY} and \textsc{RedditThreads} are social network datasets capturing interactions between users. These datasets do not originally contain node labels or attributes, for which we use node degrees as labels.

\subsection{Model Configurations}
\label{ssec:config}

For each baseline and \method, we consider a grid of various hyperparameter values. Table \ref{table:hyperparams} gives the detailed list. 

\begin{enumerate}
	\item \textbf{Proposed \method:} For our method, we vary the initialization seed, as well as number of layers, learning rate and weight decay, as these have significant effect on the performance of the model.
	For MMD-pooling, we vary our \ny subsample size - this serves two purposes: finding the ideal representation size for the training dataset, as well as helping randomize the effect of the selected \ny set. 
	We run each model for 150 epochs, where each epoch consists of mini-batch updates of size 64. 
	
	\item \textbf{Two-stage methods:} For the graph embedding/kernel techniques, we vary the hyperparameter that controls the radius of influence of each node. For PK, this is the propagation depth, whereas for WL and G2V, it is the number of iterations.
	LOF measures a point's deviation in local density from its neighbors. We vary the number of neighbors as a hyperparameter.
	For OCSVM, we set the contamination to a very small value since we are only training on inliers. On precomputed graph kernels obtained from \wl and \pk,
	there is no requirement for a bandwidth parameter. 
	For \gtovec, we use the RBF kernel with bandwidth set as per the median heuristic, i.e. $\gamma(X) = \left( \#\text{features} \cdot \text{var}(X)\right)^{-1}$.
	
	\item \textbf{OCGNN}  performs Glorot uniform weight initialization, hence we do not alter the model seed.
	We  vary the number of hidden layers and the learning rate. As suggested in the paper, we run each model for 4000 epochs.
	
	\item \textbf{DOMINANT:} For most hyperparameters, we stick to the suggested hyperparameter values from the original paper.
	$\alpha$ controls the tradeoff between the weight given to the adjacency matrix as opposed to the feature matrix of the graph.
	Additionally, we vary the learning rate and weight decay hyperparameters. As suggested in the original paper, we run each model for 300 epochs.
\end{enumerate}


\definecolor{bg}{gray}{0.85}
\begin{table}[!t]
	\centering
	\caption{Hyperparameter configs. WD: weight decay , LR: learning rate } 
	\vspace{-0.1in}
	\centering
	\scalebox{0.9}{
		\begin{tabular}{ r c c c c} 
			\hline
			\textbf{Hyperparameter/Seed} & \textbf{Lower} & \textbf{Upper} & \textbf{Step} & \#\textbf{Models}\\
			\hline
			\rowcolor{bg} \method-MMD  & & & & 162\\
			layers $L$ & $1$ & $4$ & $\times 2$  & 3\\
			WD $\lambda$ & $1 \cdot 10^{-5}$ & $1 \cdot 10^{-3}$ & $\times 10$  & 3\\
			LR $\eta$ & $0.01$ & $0.1$ & $\times 10$  & 2\\ 
			model seed & $0$ & $2$ & $+ 1$  & 3\\ 
			\ny subsample size $k$ & $4 \log n$ & $16 \log n$ & $\times 2$ & 3\\ 
			\rowcolor{bg} \method-Mean  & & & & 54\\
			layers & $1$ & $4$ & $\times 2$ & 3\\
			WD & $1 \cdot 10^{-5}$ & $1 \cdot 10^{-3}$ & $\times 10$ & 3\\
			LR & $1 \cdot 10^{-4}$ & $1 \cdot 10^{-3}$ & $\times 10$ & 2\\
			model seed & $0$ & $2$ & $+ 1$ & 3\\ 
			\midrule
			\rowcolor{bg} \pk-LOF  & & & & 25\\
			Propagation depth & $1$ & $16$ & $\times 2$ & 5\\ 
			\#neighbors (LOF) & $5$ & $80$ & $\times 2$ & 5\\ 
			\rowcolor{bg} \pk-OCSVM  & & & & 5\\
			Propagation depth & $1$ & $16$ & $\times 2$ & 5\\
			\rowcolor{bg} \wl-LOF  & & & & 25\\
			WL iterations & $1$ & $16$ & $\times 2$ & 5\\ 
			\#neighbors (LOF) & $5$ & $80$ & $\times 2$ & 5\\ 
			\rowcolor{bg} \wl-OCSVM  & & & & 5\\
			WL iterations & $1$ & $16$ & $\times 2$ & 5\\
			\rowcolor{bg} \gtovec-LOF  & & & & 75\\
			G2V iterations & $1$ & $16$ & $\times 2$ & 5\\ 
			G2V seed & $42$ & $44$ & $+ 1$ & 3\\ 
			\#neighbors (LOF) & $5$ & $80$ & $\times 2$ & 5\\ 
			\rowcolor{bg} \gtovec-OCSVM  & & & & 15\\
			G2V iterations & $1$ & $16$ & $\times 2$ & 5\\ 
			G2V seed & $42$ & $44$ & $+ 1$ & 3\\ 
			\rowcolor{bg} OCGNN  & & & & 6\\
			layers & $1$ & $4$ & $\times 2$ & 3\\ 
			LR & $0.001$ & $0.01$ & $\times 10$ & 2\\  
			\rowcolor{bg} DOMINANT  & & & & 8\\
			$\alpha$ & $0.4$ & $0.8$ & $+ 0.4$ & 2\\ 
			LR & $0.0005$ & $0.001$ & $\times 2$ & 2\\ 
			WD & $0$ & $5e$-$4$ & $+ 5e$-$4$ & 2\\ 
			
			\hline
		\end{tabular}
	}
\vspace{-0.15in}
	\label{table:hyperparams} 
\end{table}

\hide{
\subsection{Mean-pooling vs. MMD-pooling}
\label{ssec:meanvsmmd}
\la{we will probably ditch subsection this due to space...}

We analyze two types of generated anomalies, parametrized by a contamination parameter $\delta$:
\begin{itemize}
	\item Inlier distribution: 100 points, where each point is represented by a 64-sized vector of numbers sampled from $\mathcal{U}(0,1)$.
	\item Point-wise anomalies: 100 points, where each point is represented by a 64-sized vector of numbers sampled from $\mathcal{U}(0,1)$, except one point in which the numbers are instead sampled from $\mathcal{U}(\delta,1+\delta)$
	\item Distributional anomalies: 100 points, where each point is represented by a 64-sized vector of numbers sampled from $\mathcal{N}(0.5,\delta)$.
\end{itemize}

\begin{figure}[!ht]
	\begin{tabular}{cc}
	\hspace{-0.15in}	\includegraphics[width=0.55\linewidth]{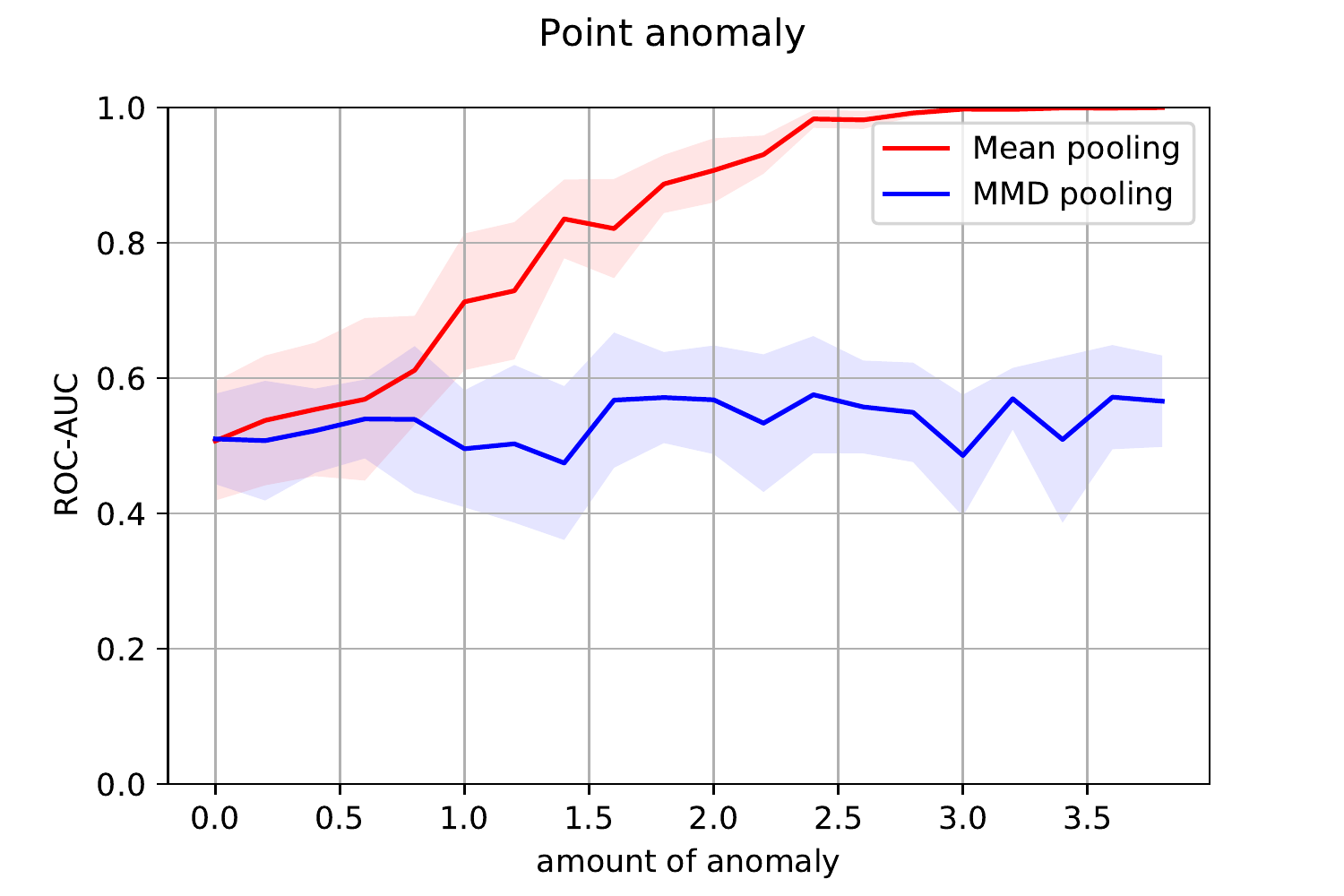}&
		\hspace{-0.25in}	\includegraphics[width=0.55\linewidth]{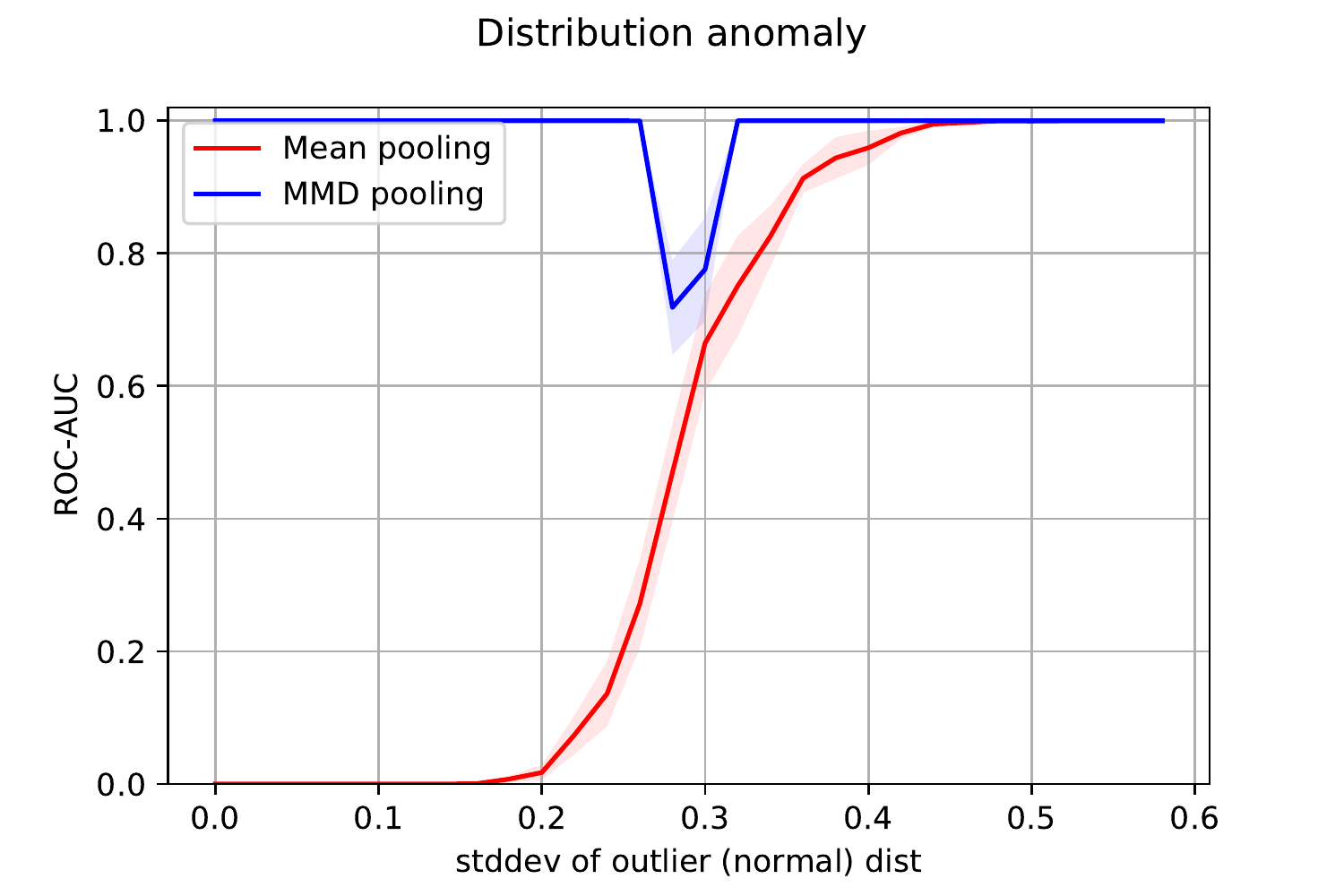}
	\end{tabular}
\vspace{-0.1in}
	\caption{Two types of anomalous sets - showing that Mean-pooling and MMD-pooling can be complementary. We run both experiments 10 times and report the mean and standard deviation of the observed ROC-AUC values. \label{fig:MMDvMean}}
	\vspace{-0.2in}
\end{figure}

From Figure~\ref{fig:MMDvMean}, we can see that Mean pooling is able to capture and identify point-wise anomalies better.
While Mean-pooling's ability to capture a point anomaly increases with the magnitude of anomaly, MMD-pooling fails to capture the difference between the two sets.

On the other hand, when the points have similar support sets but vary in the way they are distributed, MMD pooling performs better.
Mean-pooling fails to differentiate the two sets when the variance is small, but as the variance increases the support set of the uniformly distributed set starts to look different. Despite the distributional mean being the same, the sample mean begins to vary as the variance increases and the size of the dataset remains constant.
MMD pooling is able to tell apart the two distributions - however it suffers a mild dip in performance when the distributions have similar variance.
}


\hide{
	
	\begin{figure*}[!ht]
		\hspace{-0.1in}
		\includegraphics[width=0.475\linewidth]{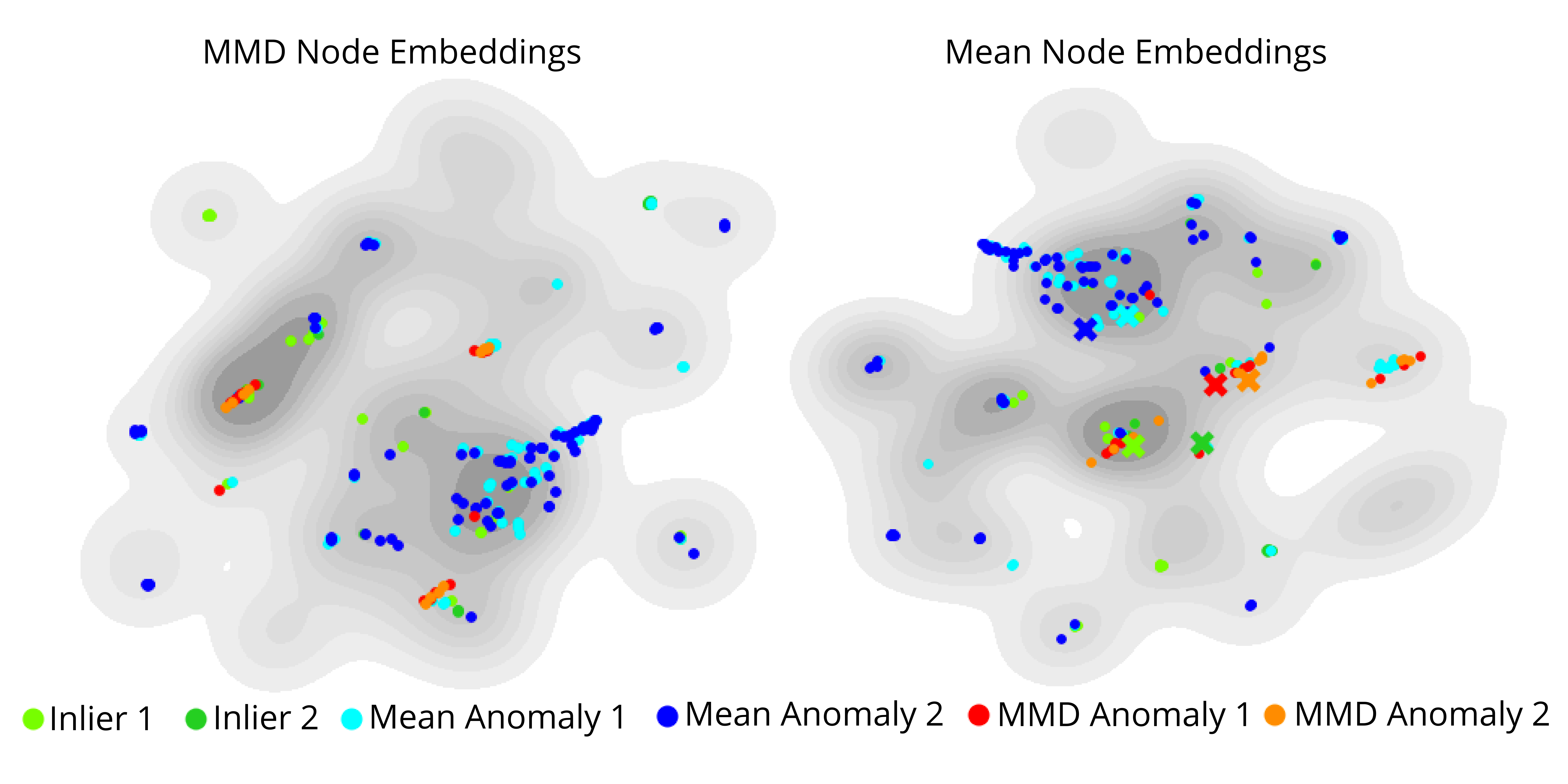}
		\hspace{-0.1in}
		\includegraphics[width=0.265\linewidth]{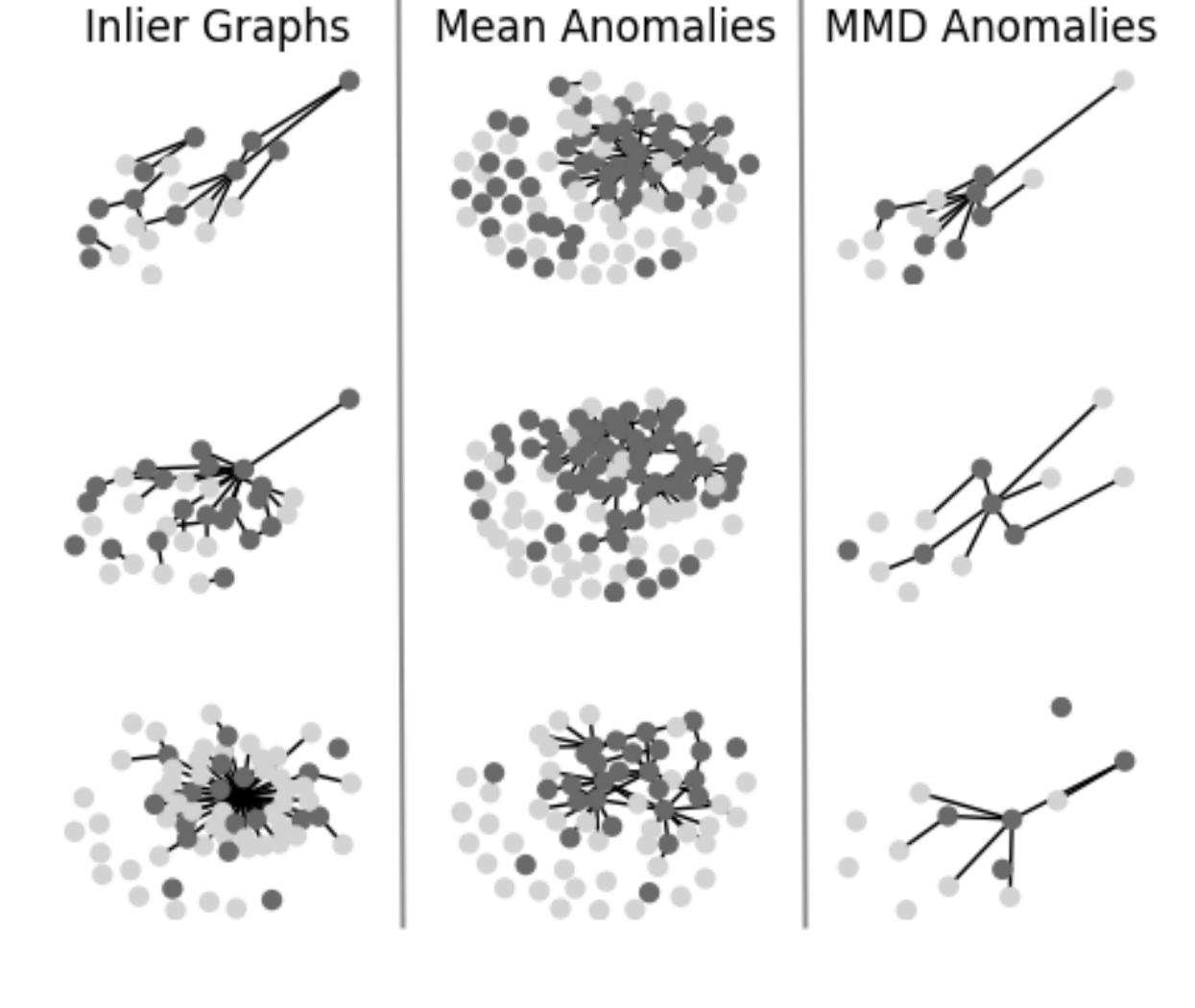}
		\includegraphics[width=0.27\linewidth]{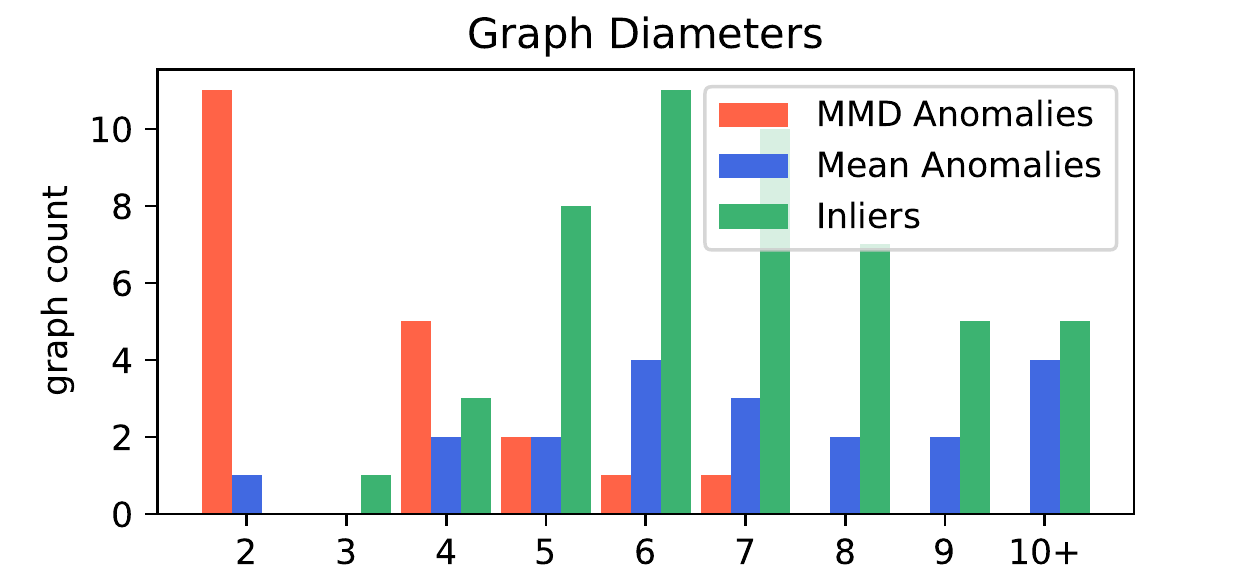}
		
		\vspace{-0.1in}
		\caption{(left) Node embedding space by MMD- and Mean-pooling in \textsc{Reddit}. Heatmap (gray) reflects the background; i.e. density of all node embeddings across $\mG$. Symbols w/ same color are nodes of the same graph (see legend). 
			Notice that the distribution of node-embeddings for MMD-anomalies differs significantly from background.
			(middle) Typical inlier graphs, and top Mean- and MMD-anomalous graphs.
			(right) Graph diameter distribution for MMD anomalies, Mean anomalies and sample inliers from \textsc{Reddit}. MMD anomalies have a distinct distribution as compared to Mean anomalies and inliers. MMD-pooling, which performs better than Mean-pooling on this dataset, is able to capture true structural anomalies which are predominantly low-diameter graphs in \textsc{Reddit}. This is also apparent from the graph visualizations (middle).
			\label{fig:case}}
		\vspace{-0.1in}
	\end{figure*}

\subsection{UMS for Disentangled Representation Learning}
\label{ssec:udrmc}

In the following, we briefly describe two existing unsupervised model selection (UMS) strategies originally proposed for unsupervised disentangled representation learning.

\subsubsection{\bf Unsupervised Disentanglement Ranking (\udr)} 
\label{ssec:udr}
\udr is ``the first method for unsupervised model selection for variational disentangled representation learning'' \cite{DuanMSWBLH20}. 
Each model is considered to be a $\{${\tt HPconfiguration}, {\tt seed}$\}$ pair.
Reciting Tolstoy who wrote ``Happy families are all alike; every unhappy family is unhappy in its own way.'', 
their main hypothesis is that
a model with a good HP setting will produce \textit{similar results} under different random initializations (i.e. seeds) whereas for a poor HP setting, results based on different random seeds will look arbitrarily different.

In a nutshell, 
\udr follows 4 steps: (1) Train $M=H\times S$ models, where 
$H$ and $S$ are the number of different HP settings 
and 
random seeds, respectively. 
(2) For each model $M_i$,  
randomly sample (without replacement) $P<S$ other models with the \textit{same HP} as $M_i$, 
but \textit{different seed}. (3) Perform $P$ pairwise comparisons between $M_i$ 
and the models sampled in Step 2 for $M_i$. (4) Aggregate pairwise similarity scores (denoted $\text{sim}_{ij}$) as
$UDR_i = \text{median}_{j}\;\text{sim}_{ij}$, for $i$$=$$1$$:$$M$ and $j$$=$$1$$:$$P$.
The model with the largest $UDR_i$ is then selected. 

Intuitively, \udr selects the model with the most stable HP setting, which yields \textit{consistent} results across various seeds.
A key part of \udr is how pairwise model comparisons are done in Step 3.
For outlier detection, we measure the {ranking similarity} of the graphs by two models using Spearman's rank-order correlation. 




\subsubsection{\bf ModelCentrality (\mc)} 

A follow-up work \cite{lin2020infogan} to \udr proposed ModelCentrality (\mc) 
for what they call ``self-supervised'' model selection for disentangling GANs. 

Their premise is similar, that ``well-disentangled models should be close to the optimal model, and hence also close to each other''.
Specifically, \mc of a model $M_i$ is given as $\mc_i = \frac{1}{M-1} \sum_{j \neq i} \text{sim}_{ij}$. The model with the largest $\mc_i$ is then selected, which coincides with the medoid of models in the pool -- hence the name \mc.
A shortcoming of  \mc is its quadratic complexity in the number of models as it requires all pairwise comparisons.

\subsection{Additional Case Study}

\label{ssec:case}

In Figure \ref{fig:case} we showcase another study on \textsc{Reddit}, similar to that illustrated in Figure \ref{fig:background} for \textsc{Proteins}. The observations are similar (also see the caption): 
\bit
\item (left) MMD-pooling based anomalies exhibit distinct node distributions, yet they go under the radar by Mean-pooling as it only considers the first moment; 
\item (middle) MMD-anomalies are strikingly dissimilar to typical inlier graphs visually, perhaps even more so than Mean-anomalies; and 
\item (right) the histograms confirm what the graph visualizations (middle) suggest, that a key contributor to the distribution (i.e. MMD) anomalies is their notably low diameter.
\eit

}

\end{document}